\documentclass[12pt]{article}

\usepackage[noadjust]{cite}
\usepackage{amsmath,amssymb,amsthm,authblk,color,graphicx,hyperref,xifthen}
\usepackage[margin=1in]{geometry}

\usepackage{algorithm,algcompatible}

\title{BLADE: Filter Learning for General Purpose Computational Photography}
\author{Pascal Getreuer}
\author{Ignacio Garcia-Dorado}
\author{John Isidoro}
\author{Sungjoon Choi}
\author{Frank Ong}
\author{Peyman Milanfar}
\affil{Google Research, Mountain View CA, USA}

\hypersetup{
    pdftitle={BLADE: Filter Learning for General Purpose Computational
              Photography},
    pdfauthor={Pascal Getreuer, Ignacio Garcia-Dorado,
               John Isidoro, Sungjoon Choi, Frank Ong, Peyman Milanfar},
    pdfnewwindow=true,
    colorlinks=true,
    linkcolor=red,
    citecolor=blue,
    filecolor=blue,
    urlcolor=blue
}

\usepackage{xcolor}
\definecolor{PlotColorA}{HTML}{204A87}
\definecolor{PlotColorB}{HTML}{677F34}
\definecolor{PlotColorC}{HTML}{C24C5A}

\definecolor{ColorOri00}{HTML}{EE55F2}
\definecolor{ColorOri01}{HTML}{FB83CD}
\definecolor{ColorOri02}{HTML}{FAB0A4}
\definecolor{ColorOri03}{HTML}{FED472}
\definecolor{ColorOri04}{HTML}{EFEC36}
\definecolor{ColorOri05}{HTML}{C8E816}
\definecolor{ColorOri06}{HTML}{9AD20F}
\definecolor{ColorOri07}{HTML}{5EC30D}
\definecolor{ColorOri08}{HTML}{36AE26}
\definecolor{ColorOri09}{HTML}{3E8F62}
\definecolor{ColorOri10}{HTML}{41759B}
\definecolor{ColorOri11}{HTML}{274FC5}
\definecolor{ColorOri12}{HTML}{2E24E9}
\definecolor{ColorOri13}{HTML}{6021F7}
\definecolor{ColorOri14}{HTML}{9039F8}
\definecolor{ColorOri15}{HTML}{C043FD}

\usepackage{tikz}
\pgfrealjobname{article}
\usetikzlibrary{arrows,fadings,patterns,shapes}
\usetikzlibrary{decorations.pathmorphing}
\usetikzlibrary{decorations.pathreplacing}

\renewcommand{\div}{\operatorname{div}}
\newcommand{\grad}{\nabla}
\newcommand{\units}[1]{{}\hspace*{0.2em}\ensuremath{\mathrm{#1}}}

\newcommand{\vv}[1]{\mathbf{#1}}

\newcommand{\filter}{h}
\newcommand{\inimage}{z}
\newcommand{\outimage}{\Hat{u}}
\newcommand{\targetimage}{u}

\newcommand{\comment}[1]{\textcolor{blue}{[\small{}#1]}}
\newcommand{\todo}[2][]{\comment{%
\textsc{Todo}\ifthenelse{\equal{#1}{}}{}{(#1)}: #2}}

\begin{document}

\maketitle

\centerline{\begin{minipage}{0.85 \textwidth}
\paragraph{Abstract}
The Rapid and Accurate Image Super Resolution (RAISR) method of Romano, Isidoro,
and Milanfar is a computationally efficient image upscaling method using a
trained set of filters. We describe a generalization of RAISR, which we name
Best Linear Adaptive Enhancement (BLADE). This approach is a trainable
edge-adaptive filtering framework that is general, simple, computationally
efficient, and useful for a wide range of problems in computational photography.
We show applications to operations which may appear in a camera pipeline
including denoising, demosaicing, and stylization.
\end{minipage}}

\section{Introduction}

In recent years, many works in image processing
have been based on nonlocal patch modeling. These include nonlocal
means of Buades, Coll, and Morel~\cite{buades2005nonlocal} and the
BM3D denoising method of Dabov et al.~\cite{dabov2007image} and their extensions
to other problems including deconvolution~\cite{lou2010image} and
demosaicing~\cite{dabov2008image,duran2014self}. While these methods can achieve
state-of-the-art quality, they tend to be prohibitively computationally
expensive, limiting their practical use.

Deep learning has also become popular in image processing. These methods can
trade quality vs.\ computation time and memory costs through considered choice
of network architecture. Particularly, quite a few works take inspiration from
partial differential equation (PDE) techniques and closely-connected areas of
variational models, Markov random fields, and maximum a posteriori estimation.
Roth and Black's fields of experts~\cite{roth2009fields}, among other
methods~\cite{zhu1997prior,scharr2003image,schmidt2014shrinkage}, develops forms
of penalty functions (image priors) that are trainable. Schmidt and Roth's
cascade of shrinkage fields~\cite{schmidt2014shrinkage} and Chen and Pock's
trainable nonlinear reaction diffusion~\cite{chen2017trainable} are designed as
unrolled versions of variational optimization methods, with each gradient
descent step or PDE diffusion step interpreted as a network layer, then
substituting portions of these steps with trainable variables. These hybrid deep
learning/energy optimization approaches achieve impressive results with reduced
computation cost and number of parameters compared to more generic structures
like multilayer convolutional networks~\cite{chen2017trainable}.

Deep networks are hard to analyze, however, which makes
failures challenging to diagnose and fix. Despite efforts to understand their
properties~\cite{bengio2009learning,
montavon2011kernel,mallat2016understanding,wang2016analysis}, the
representations deep networks learn and what makes them effective remain
powerful but without a complete mathematical analysis. Additionally, the
cost of running inference for deep networks is nontrivial or
infeasible for some applications on mobile devices. Smartphones
lack the computation power to do much processing in a timely fashion at
full-resolution on the photographs they capture (often +10-megapixel resolution
as of 2017). The difficulties are even worse for on-device video processing.

These problems motivate us to take a lightweight, yet
effective approach that is \emph{trainable} but still \emph{computationally
simple} and \emph{interpretable}. We describe an approach that extends the Rapid
and Accurate Image Super Resolution (RAISR) image upscaling method of Romano,
Isidoro, and Milanfar~\cite{romano2017raisr} and
others~\cite{choi2017single,jiang2017learning} to a generic method
for trainable image processing, interpreting it as a local version of
optimal linear estimation.
Our approach can be seen as a shallow two-layer
network, where the first layer is predetermined and the second layer is trained.
We show that this simple network structure allows for inference that is
computationally efficient, easy to train, interpret, and sufficiently flexible
to perform well on a wide range of tasks.

\subsection{Related work}

RAISR image upscaling~\cite{romano2017raisr} begins by computing the $2\times 2$
image structure tensor on the input low-resolution image. Then for each output
pixel location, features derived from the structure tensor are used to select a
linear filter from a set of a few thousand trained filters to compute the output
upscaled pixel. A similar upscaling method is
global regression based on local linear mappings super-interpolation (GLM-SI)
by Choi and Kim~\cite{choi2017single}, which like RAISR, upscales using trained
linear filters, but using overlapping patches that are blended in
a linearly optimal way.
The $L^3$ demosaicing method~\cite{jiang2017learning} is a
similar idea but applied only to demosaicing.
$L^3$ computes several local features, including the patch mean, variance, and
saturation, then for each pixel uses these features to select a linear
filter\footnote{More precisely, $L^3$'s trained filters are affine, they include
a bias term.} from among a trained set of filters to estimate the demosaiced
pixel. In both RAISR and $L^3$, processing is locally adaptive due to the
nonlinearity in filter selection. We show that this general approach extends
well to other image processing tasks.

Quite a few other works use collections of trainable filters. For instance,
Gelfand and Ravishankar~\cite{gelfand1993tree} consider a tree where each node
contains a filter, where in nonterminal nodes, the filter plus a threshold is
used to decide which child branch to traverse. Fanello et
al.~\cite{fanello2014filter} train a random forest with optimal linear filters
at the leaves and split nodes to decide which filter to use. Schulter, Leistner,
and Bischof~\cite{schulter2015fast} expands on this work by replacing the linear
filters with polynomials of the neighboring pixels.

Besides $L^3$, probably the closest existing work to ours is the image
restoration method of Stephanakis, Stamou, and
Kollias~\cite{stephanakis1999piecewise}. The authors use local wavelet features
to make a fuzzy partition of the image into five regions (one region describing
smooth pixels, and four regions for kinds of details). A Wiener filter is
trained for each region. During inference, each Wiener filter is applied and
combined in a weighted sum according to the fuzzy partition.

The K-LLD method of Chatterjee and Milanfar~\cite{chatterjee2009clustering} is
closely related, where image patches with similar local geometric features are
clustered and a least-squares optimal filter is learned for each cluster. The
piecewise linear estimator of Yu, Sapiro, and Mallat~\cite{yu2012solving}
similarly uses an E-M algorithm that alternatingly clusters patches of the image
and learns a filter for each cluster. Our work can be seen as a simplification
of these methods using predetermined clusters.

\subsection{Contribution of this work}

We extend RAISR~\cite{romano2017raisr} to a trainable filter framework for
general imaging tasks, which we call Best Linear Adaptive Enhancement (BLADE).
We interpret this extension as a spatially-varying filter based on a locally
linear estimator.

In contrast to Stephanakis~\cite{stephanakis1999piecewise}, we make a hard
decision at each pixel of which filter to apply, rather than a soft (fuzzy) one.
Unlike Yu et al.~\cite{yu2012solving}, our filter selection step is a simple
uniform quantization, avoiding the complication of a general nearest cluster
search. Notably, these properties make our computation cost independent of the
number of filters, which allows us to use many filters (often hundreds
or thousands) while maintaining a fast practical system.

Specifically, unique attributes of our approach are:
\begin{itemize}
\item Inference is very fast, executing in real-time on a typical mobile
device. Our CPU implementation for $7\times 7$ filters produces
22.41\units{MP/s} on a 2017~Pixel phone.
\item Training is solvable by basic linear algebra, where training of each
filter amounts to a multivariate linear regression problem.
\item The approach is sufficiently flexible to perform well on a range of
imaging tasks.
\item Behavior of the method is interpretable in terms of the set of
trained linear filters, including diagnostics to catch problems in training.
\end{itemize}

\subsection{Outline}

We introduce BLADE in section~\ref{sec:blade}. Filter selection based on the
image structure tensor is described in section~\ref{sec:filter_selection}.
Sections~\ref{e:learning_image_processing_filters} and
sections~\ref{sec:denoising}, \ref{sec:jpeg_artifact_removal}, and 
\ref{sec:demosaicing} demonstrate several applications
of BLADE. Computational performance at
inference is discussed in section~\ref{sec:computational_performance}.
Section~\ref{sec:conclusions} concludes the paper.

\begin{figure}[t]
\centering
\mbox{%
\beginpgfgraphicnamed{images/inference_diagram}%

\begin{tikzpicture}[scale=0.45,>=stealth',semithick]

\begin{scope}
\node [rectangle] (input) at (0,0) {{\small input patch}
$\vv{R}_i \vv{\inimage}$};
\coordinate (split0) at (0,1.5) {};

\begin{scope}[yshift=7.2cm]
\node [rectangle,draw,inner sep=3pt] (h1) at (-3.35,0)
{\small $\vv{\filter}^1$};
\node [rectangle,draw,inner sep=3pt] (h2) at (-0.85,0)
{\small $\vv{\filter}^2$};
\node (h3) at (1.25,0) {\small $\cdots$};
\node [rectangle,draw,inner sep=3pt] (h4) at (3.35,0)
{\small $\vv{\filter}^K$};
\draw (h1) node [left=1.5em] {\small linear filterbank};
\end{scope}

\node [rectangle,draw,inner sep=3pt] (stanalysis) at (4,2.9)
{\small $s(i)$};
\draw (stanalysis) node [right=2em] {\small filter selection};

\node [rectangle] (output) at (0,11) {{\small output pixel} $\outimage_i$};

\begin{scope}[yshift=8.5cm]
\coordinate (split) at (0,0) {};
\draw (split) -- (split-|h1) to (h1);
\draw (split-|h2) to (h2);
\draw (split) -- +(0.6,0);
\draw [dashed] (0.6,0) -- (1.6,0);
\draw (1.6,0) -- (split-|h4) -- (h4);
\draw [->] (split) -- (output);
\end{scope}

\draw [->] (split0) -- (split0-|stanalysis) -- (stanalysis);

\begin{scope}[yshift=6cm]
\node [circle,draw,inner sep=1pt] (h1c) at (-3.35,0) {};
\node [circle,draw,inner sep=1pt] (h2c) at (-0.85,0) {};
\node [circle,draw,inner sep=1pt] (h4c) at (3.35,0) {};
\end{scope}

\node [circle,draw,inner sep=1pt] (sel) at (0,4) {};

\draw (h1) to (h1c);
\draw (h2) to (h2c);
\draw (h4) to (h4c);

\draw [xshift=0cm,yshift=4cm,very thick,gray!50] (40:1.2) arc (40:140:1.2);
\draw [shorten >=4pt] (sel) -- (h1c);
\draw [->,shorten >=2pt] (stanalysis) -- (stanalysis|-sel) -- (sel);
\end{scope}

\draw (input) -- (sel);


\end{tikzpicture}%
\endpgfgraphicnamed}
\caption{\label{fig:inference_diagram} BLADE inference.
$\vv{R}_i$ denotes extraction of a patch centered at pixel $i$. For a given
output pixel $\outimage_i$, we only need to evaluate the one linear filter that
is selected, $\filter^{s(i)}$.}
\end{figure}
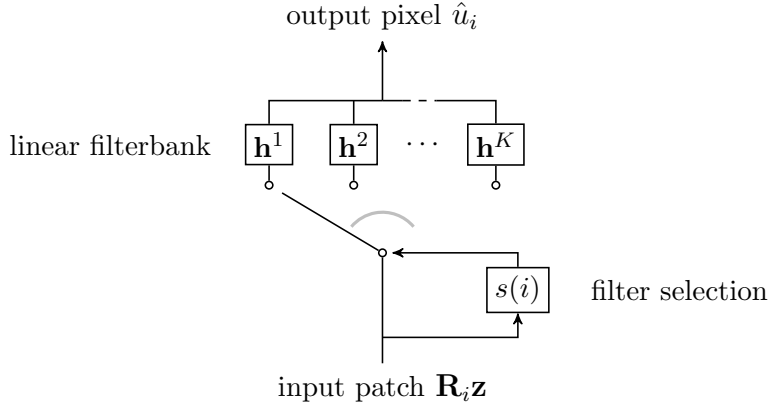

\section{Best Linear Adaptive Enhancement}\label{sec:blade}

This section introduces our Best Linear Adaptive Enhancement (BLADE) extension
of RAISR.

We denote the input image by $\vv{\inimage}$ and subscripting $\inimage_i$
to denote the value at spatial position $i\in\Omega\subset\mathbb{Z}^2$. Let
$\vv{\filter}^1,\ldots,\vv{\filter}^K$ be
a set of linear FIR filters, where superscript indexes different filters and
each filter has nonzero support or footprint $F\subset\mathbb{Z}^2$. Inference
is a spatially-varying correlation. The essential idea of RAISR is that for
inference of each output pixel $\outimage_i$, one filter in the bank is
selected:
\begin{equation}\label{e:blade_inference}
\outimage_i = \sum_{j\in F} \filter^{s(i)}_j \, \inimage_{i + j},
\end{equation}
where $s(i) \in \{1,\ldots,K\}$ selects the filter applied at spatial position
$i$. This spatially-adaptive filtering is equivalent to passing $\vv{\inimage}$
through a linear filterbank and then for each spatial position selecting one
filter output (Figure~\ref{fig:inference_diagram}). We stress that in
computation, however, we only evaluate the one linear filter that is selected.
Both the selection $s$ and spatially-varying filtering (\ref{e:blade_inference})
are vectorization and parallelization-friendly so that inference can be
performed with high efficiency.

Let $N = |F|$ be the filter footprint size. The number of arithmetic operations
per output pixel in (\ref{e:blade_inference}) is $O(N)$, proportional to the
footprint size, independent of the number of filters $K$ because we make a hard
decision to select one filter at each pixel location. A large number of filters
may be used without impacting computation time, so long as the filters fit in
memory and filter selection $s$ is efficient.

To make filtering adaptive to edges and local structure, we perform filter
selection $s$ using features of the $2\times 2$ image structure tensor
(discussed in section~\ref{sec:filter_selection}). We determine the filter
coefficients by training on a dataset of observed and target example pairs,
using a simple $L^2$ loss plus a quadratic regularizing penalty. This
optimization decouples such that the filters are individually solvable in closed
form, in which each filter amounts to a regularized multivariate linear
regression~\cite{romano2017raisr}.

\begin{figure}[t]
\centering
\mbox{%
\beginpgfgraphicnamed{images/subsets}%
\input{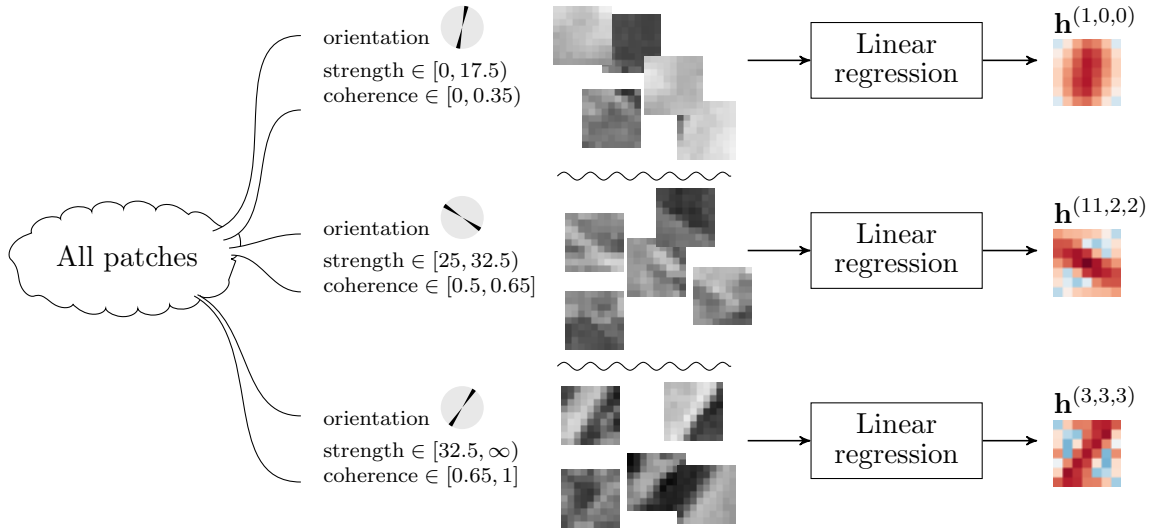}%
\endpgfgraphicnamed}
\caption{\label{fig:subsets} BLADE training. Linear filters are trained
piecewise on different
subsets of patches. Filter selection by local orientation, strength, and
coherence (explained in section~\ref{sec:filter_selection}) partitions input
patches into multiple subsets, and we regress a linear filter on each subset.}
\end{figure}

\subsection{Inference}

Rewriting inference (\ref{e:blade_inference}) generically,\footnote{Yet more
generically, signals could be vector-valued or on domains of other dimension,
though we will focus on two-dimensional images.} each output pixel
is an inner product between some selected filter and patch extracted from the
input,
\begin{equation}
\outimage_i = (\vv{\filter}^{s(i)})^T \vv{R}_i \vv{\inimage},
\end{equation}
where $(\cdot)^T$ denotes matrix transpose and $\vv{R}_i$ is an operator that
extracts a patch centered at $i$, $(\vv{R}_i \vv{\inimage})_j = \inimage_{i+j}$,
$j\in F$. The operation over the full image can be seen to be a matrix-vector
multiplication,
\begin{equation}
\vv{\outimage} = \vv{W} \vv{\inimage},
\end{equation}
where $\vv{W}$ is a data-dependent matrix in which the $i$th row of $\vv{W}$ is
$\vv{R}_i^T \vv{\filter}^{s(i)}$.

\subsection{Training}

We use a quadratic penalty for filter regularization. Given an $N \times N$
nonnegative definite matrix $\vv{Q}$, define the seminorm for a filter
$\vv{\filter}$
\begin{equation}
\|\vv{\filter}\|_\vv{Q} := (\vv{\filter}^T \vv{Q} \vv{\filter})^{1/2}.
\end{equation}
For most applications, we set the regularizing $\vv{Q}$ matrices such that
$\|\vv{\filter}\|_\vv{Q}$ is a discretization of $L^2$ norm of the filter's
spatial gradient (also known as the $H^1$ or Sobolev $W^{1,2}$ seminorm),
\begin{equation}
\| \vv{\filter} \|_Q^2 = \frac{\lambda}{2} \sum_{\substack{
i,j\in F: \\ \|i-j\| = 1}}|\filter_i - \filter_j|^2
\end{equation}
where $\lambda$ is a positive parameter controlling the regularization strength.
This encourages filters to be spatially smooth.

Given an observed input image $\vv{\inimage}$ and corresponding target image
$\vv{\targetimage}$, we train the filter coefficients as the solution of
\begin{equation}\label{e:training_objective}
\operatorname*{arg\,min}_{\vv{\filter}^1,\ldots,\vv{\filter}^K} \,
    \sum_{k=1}^K \|\vv{\filter}^k\|_\vv{Q}^2
    + \|\vv{\targetimage} - \vv{\outimage}\|^2,
\end{equation}
in which, as above, $\outimage_i = (\vv{\filter}^{s(i)})^T \vv{R}_i \vv{z}$. The
objective function can be decomposed as
\begin{equation}
\sum_{k=1}^K \Bigl(\|\vv{\filter}^k\|_\vv{Q}^2
    + \sum_{\substack{i\in\Omega:\\ s(i)=k}}
    |\outimage_i - (\vv{\filter}^k)^T \vv{R}_i \vv{z}|^2 \Bigr)
\end{equation}
so that the minimization decouples over $k$, where the inner sum is over the
subset of pixel locations
where filter $\vv{\filter}^k$ is selected. This means we can solve for each
filter independently. Filter selection effectively
partitions the training data into $K$ disjoint subsets, over which each filter
is trained. Figure~\ref{fig:subsets} shows an
example of what these patch subsets look like.

Training from multiple such observed/target image pairs is similar. We include
spatial axial flips and $90^\circ$ rotations of each observation/target image
pair in the training set (effectively multiplying the amount of training data by
a factor of 8, for ``free'') so that the filters learn symmetries with respect
to these manipulations.

The subproblem for each filter $\vv{\filter}$ takes the following form: let
$\{i(1),\ldots,i(M)\}$ enumerate spatial positions where $s(i) = k$ and $M$
is the number of such locations, and
define $b_m = x_{i(m)}$ and $\Hat{b}_m = \vv{\filter}^T \vv{R}_{i(m)} \vv{z}$.
We solve for the filter as
\begin{equation}
\operatorname*{arg\,min}_{\vv{\filter}} \, \|\vv{\filter}\|_\vv{Q}^2
    + \|\vv{b} - \vv{\Hat{b}}\|^2.
\end{equation}
This is simply multivariate linear regression with regularization. We review it
briefly here. Defining the design matrix $A_{m,n} = (\vv{R}_m z)_n$ of size
$M \times N$, the squared residual norm is
\begin{align}
\| \vv{b} - \vv{\Hat{b}}\|^2
&= \sum_{m=1}^M
| b_m - \Hat{b}_m |^2 \nonumber \\
&= \sum_{m=1}^M
\Bigl| b_m - \sum_{n=1}^N A_{m,n} \filter_n \Bigr|^2 \nonumber \\
&= \|\vv{b}\|^2 - 2 \vv{A}^T \vv{b} + \|\vv{\filter}\|_{\vv{A}^T \vv{A}}^2.
\end{align}
The optimal filter is
\begin{equation}
\vv{\filter} = (\vv{Q} + \vv{A}^T \vv{A})^{-1} \vv{A}^T \vv{b}.
\end{equation}

\begin{algorithm}
\caption{BLADE training}
\begin{algorithmic}[1]
\REQUIRE Observed image $\vv{\inimage}$ and target image $\vv{\targetimage}$
\ENSURE Filter $\vv{\filter}^k$,
residual variance $\sigma_\vv{r}^2$, filter variance estimate
$\Sigma_{\vv{\filter}^k}$
\STATE Determine filter selection $s$.
\STATE Initialize $\vv{G}$ and $M$ with zeros.
\FOR{each $i$ where $s(i) = k$}
  \STATE $\vv{G} \leftarrow \vv{G} + \bigl(
  \begin{smallmatrix}\vv{R}_i \vv{\inimage} \\ x_i\end{smallmatrix}\bigr)
  (\!\begin{smallmatrix}\vv{R}_i \vv{\inimage}^T & x_i\end{smallmatrix}\!)$
  \STATE $M \leftarrow M + 1$
\ENDFOR
\STATE $
    \bigl(\begin{smallmatrix}
    \vv{A}^T \vv{A} & \vv{A}^T \vv{b} \\
    \vv{b}^T \vv{A} & \vv{b}^T \vv{b}
    \end{smallmatrix}\bigr) \leftarrow \vv{G}$
\STATE $\vv{\filter}^k \leftarrow (\vv{Q}
    + \vv{A}^T \vv{A})^{-1} \vv{A}^T \vv{b}$
\STATE $\sigma_\vv{r}^2 \leftarrow
    \tfrac{1}{M - N} (\vv{b}^T \vv{b} - 2 \vv{A}^T \vv{b}
    + \|\vv{\filter}^k\|_{\vv{A}^T \vv{A}}^2)$
\STATE $\Sigma_{\vv{\filter}^k} \leftarrow
    \sigma_\vv{r}^2 (\vv{Q} + \vv{A}^T \vv{A})^{-1}$
\end{algorithmic}
\end{algorithm}

The variance of the residual $\vv{r} = \vv{b} - \vv{\Hat{b}}$ is estimated as
\begin{equation}
\sigma_\vv{r}^2 = \tfrac{1}{M - N} \|\vv{b} - \vv{\Hat{b}}\|^2.
\end{equation}
Modeling the residual as i.i.d.\ zero-mean normally distributed noise of
variance $\sigma_\vv{r}^2$, an estimate of the filter's covariance matrix is
\begin{equation}
\Sigma_\vv{\filter} = \sigma_\vv{r}^2 (\vv{Q} + \vv{A}^T \vv{A})^{-1}.
\end{equation}
The square root of the $j$th diagonal element estimates the standard deviation
of $\filter_j$, which is a useful indication of its reliability. In
implementation, it is enough to accumulate a Gram matrix of size $(N+1)$ by
$(N+1)$, which allows us to train from any number of examples with a fixed
amount of memory. The above algorithm can be used to train multiple filters in
parallel.

RAISR~\cite{romano2017raisr} is a special case of BLADE where the observations
$\vv{\inimage}$ are downscaled versions of the targets $\vv{\targetimage}$ and
a different set of filters are learned for each subpixel shift.

In the extreme $K=1$ of a single filter, the trained filter is the classic
Wiener filter~\cite{wiener1949extrapolation}, the linear minimum mean square
estimator relating the observed image to the target. With multiple filters $K >
1$, the result is necessarily at least as good in terms of MSE as the Wiener
filter. Since filters are trained over different subsets of the data, each
filter is specialized for its own distribution of input patches. This
distribution may be a much narrower and more accurately modeled than the
one-size-fits-all single filter estimator.

BLADE may be seen as a particular two-layer neural network architecture, where
filter selection $s(i)$ is the first layer and filtering with $\vv{h}^{s(i)}$ is
the second layer (Figure~\ref{fig:inference_diagram}). An essential feature is
that BLADE makes a hard decision in which filter to apply. Hard decision is
usually avoided intermediately in a neural network since it is necessarily a
discontinuous operation, for which gradient-based optimization techniques are
not directly applicable.

A conventional network architecture for edge-adaptive filtering
would be to use a convolutional neural network (CNN) in which later layers make
weighted averages of filtered channels from the previous layer, essentially
a soft decision among filters as described for example by Xu et
al.~\cite{xu2015deep}. Soft decisions are differentiable and
more easily trainable, but requires in a CNN that all filters are evaluated at
all spatial positions, so cost increases with the number of filters. On the
contrary, BLADE's inference cost is independent of the number of filters, since
for each spatial position only the selected filter is evaluated.

Besides efficient inference, a strength of our approach is interpretability of
the trained filters. It is possible to plot the table of filters (examples are
shown in Sections~\ref{e:learning_image_processing_filters},
\ref{sec:denoising}, \ref{sec:demosaicing}) and assess the behavior by visual
inspection. Defects can be quickly identified, such as inadequate training data
or regularization manifest as filters with noisy coefficients, and using an
unnecessarily wide support is revealed by all filters having a border of noisy
or small coefficients.

\section{Filter Selection}\label{sec:filter_selection}

To make filtering (\ref{e:blade_inference}) adaptive to image edges and local
structure, we perform filter selection $s$ using features of the $2\times 2$
image structure tensor. Ideally, filter selection should partition the input
data finely and uniformly enough that each piece of the input data over
$\{i\in\Omega : s(i)=k\}$ is well-approximated by a linear estimator and
contains an adequate number of training examples. Additionally, filter selection
should be robust to noise and computationally efficient.

We find that structure tensor analysis is an especially good choice: it is
robust, reasonably efficient, and works well for a range of tasks that we have
tested. The structure tensor analysis is a principle components analysis (PCA)
of the local gradients. PCA explains the variation in the gradients along the
principal directions. In a small window, one can argue this is all that matters
to understand the geometry of the signal. Generically though, any features
derived from the input image could be used. For example, the input intensity
could be used in filter selection to process light vs.\ dark regions
differently.

\subsection{Image structure tensor}

As introduced by F\"orstner and G\"ulch~\cite{forstner1987fast} and Big\"un and
Granlund~\cite{bigun1987optimal}, the image structure tensor (also known as the
second-moment matrix, scatter matrix, or interest operator) is
\begin{equation}\label{e:structure_tensor}
J(\grad u) :=
\left(\begin{matrix} \partial_{x_1} u \\ \partial_{x_2} u
\end{matrix}\right)
\left(\begin{matrix} \partial_{x_1} u & \partial_{x_2} u
\end{matrix}\right),
\end{equation}
where in the above formula, $u(x)$ is a continuous-domain image,
$\partial_{x_1}$ and $\partial_{x_2}$ are the spatial partial derivatives in
each axis, and $\grad = (\partial_{x_1}, \partial_{x_2})^T$ denotes gradient.
At each point, $J(\grad u)(x)$ is a $2\times 2$ rank-1 matrix formed
as the outer product of the gradient $\grad u(x)$ with itself. The structure
tensor is smoothed by convolution with Gaussian $G_\rho$ of standard deviation
$\rho$,
\begin{equation}
J_\rho(\grad u) := G_\rho * J(\grad u),
\end{equation}
where the convolution is applied spatially to each component of the tensor. The
smoothing parameter $\rho$ determines the scale of the analysis, a larger
$\rho$ characterizes the image structure over a larger neighborhood.

As described e.g.\ by Weickert~\cite{weickert1998anisotropic}, the $2\times 2$
matrix at each pixel location of the smoothed structure tensor $J_\rho(\grad u)$
is nonnegative definite with orthogonal eigenvectors. This eigensystem is a
powerful characterization of the local image geometry. The dominant eigenvector
is a robust spatially-smoothed estimate of the gradient orientation (the
direction up to a sign ambiguity) while the second eigenvector gives
the local edge orientation. The larger eigenvalue is a smoothed estimate
of the squared gradient magnitude. In other words, the eigensystem of
$J_\rho(\grad u)$ is a spatially-weighted principle components analysis of the
raw image gradient $\grad u$.

The eigenvalues $\lambda_1 \ge \lambda_2$ and dominant eigenvector $\vv{w}$ of a
matrix $\bigl(\begin{smallmatrix} a & b \\ b & c\end{smallmatrix}\bigr)$ can be
computed as
\begin{align}
\lambda_{1,2} &= \tfrac{1}{2} \bigl((a + c) \pm \delta\bigr), \\
\vv{w} &= \left(\begin{matrix}
2 b \\ c - a + \delta
\end{matrix}\right), \label{e:v1_computation}
\end{align}
where\footnote{If the matrix is proportional to identity ($a = c$, $b = 0$), any
vector is an eigenvector of the same eigenvalue. In this edge case,
(\ref{e:v1_computation}) computes $\vv{w} = \bigl(\begin{smallmatrix}0 \\
0\end{smallmatrix}\bigr)$, which might be preferable as an isotropic
characterization.} $\delta = \sqrt{(a - c)^2 + (2 b)^2}$. The second eigenvector
is the orthogonal complement of $w$. From the eigensystem, we define the
features:
\begin{itemize}
\item $\text{orientation} = \arctan w_2/w_1$, is the
predominant local orientation of the gradient;
\item $\text{strength} = \sqrt{\lambda_1}$, is the local gradient magnitude; and
\item $\text{coherence} = \frac{\sqrt{\lambda_1} - \sqrt{\lambda_2}}{
\sqrt{\lambda_1} + \sqrt{\lambda_2}}$, which characterizes the amount of
anisotropy in the local structure.\footnote{Different works vary in the details
of how ``coherence'' is defined. This is the definition that we use.}
\end{itemize}

\begin{figure}[t]
\centering
\mbox{%
\beginpgfgraphicnamed{images/quantization}%

\begin{tikzpicture}[scale=1.4]

\begin{scope}
\fill [ColorOri00!40] ((0:0) -- (84.375:1) arc (84.375:95.625:1)
  -- (95.625:-1) arc (95.625:84.375:-1) -- cycle;
\fill [ColorOri01!40] ((0:0) -- (73.125:1) arc (73.125:84.375:1)
  -- (84.375:-1) arc (84.375:73.125:-1) -- cycle;
\fill [ColorOri02!40] ((0:0) -- (61.875:1) arc (61.875:73.125:1)
  -- (73.125:-1) arc (73.125:61.875:-1) -- cycle;
\fill [ColorOri03!40] ((0:0) -- (50.625:1) arc (50.625:61.875:1)
  -- (61.875:-1) arc (61.875:50.625:-1) -- cycle;
\fill [ColorOri04!40] ((0:0) -- (39.375:1) arc (39.375:50.625:1)
  -- (50.625:-1) arc (50.625:39.375:-1) -- cycle;
\fill [ColorOri05!40] ((0:0) -- (28.125:1) arc (28.125:39.375:1)
  -- (39.375:-1) arc (39.375:28.125:-1) -- cycle;
\fill [ColorOri06!40] ((0:0) -- (16.875:1) arc (16.875:28.125:1)
  -- (28.125:-1) arc (28.125:16.875:-1) -- cycle;
\fill [ColorOri07!40] ((0:0) -- (5.625:1) arc (5.625:16.875:1)
  -- (16.875:-1) arc (16.875:5.625:-1) -- cycle;
\fill [ColorOri08!40] ((0:0) -- (-5.625:1) arc (-5.625:5.625:1)
  -- (5.625:-1) arc (5.625:-5.625:-1) -- cycle;
\fill [ColorOri09!40] ((0:0) -- (-16.875:1) arc (-16.875:-5.625:1)
  -- (-5.625:-1) arc (-5.625:-16.875:-1) -- cycle;
\fill [ColorOri10!40] ((0:0) -- (-28.125:1) arc (-28.125:-16.875:1)
  -- (-16.875:-1) arc (-16.875:-28.125:-1) -- cycle;
\fill [ColorOri11!40] ((0:0) -- (-39.375:1) arc (-39.375:-28.125:1)
  -- (-28.125:-1) arc (-28.125:-39.375:-1) -- cycle;
\fill [ColorOri12!40] ((0:0) -- (-50.625:1) arc (-50.625:-39.375:1)
  -- (-39.375:-1) arc (-39.375:-50.625:-1) -- cycle;
\fill [ColorOri13!40] ((0:0) -- (-61.875:1) arc (-61.875:-50.625:1)
  -- (-50.625:-1) arc (-50.625:-61.875:-1) -- cycle;
\fill [ColorOri14!40] ((0:0) -- (-73.125:1) arc (-73.125:-61.875:1)
  -- (-61.875:-1) arc (-61.875:-73.125:-1) -- cycle;
\fill [ColorOri15!40] ((0:0) -- (-84.375:1) arc (-84.375:-73.125:1)
  -- (-73.125:-1) arc (-73.125:-84.375:-1) -- cycle;

\draw (0,0) circle (1);

\foreach \th in {-84.375, -73.125, -61.875, -50.625, -39.375, -28.125, -16.875,
-5.625, 5.625, 16.875, 28.125, 39.375, 50.625, 61.875, 73.125, 84.375}
{
  \draw (\th:-1) -- (\th:1);
}
\foreach \i/\th in {0/90, 1/78.75, 2/67.5, 3/56.25, 4/45, 5/33.75, 6/22.5,
7/11.25, 8/-0, 9/-11.25, 10/-22.5, 11/-33.75, 12/-45, 13/-56.25, 14/-67.5,
15/-78.75}
{
  \draw (\th:1.15) node {\tiny $\i$};
}

\draw (0,-1.6) node {\small Orientation\vphantom{Sg}};
\end{scope}

\begin{scope}[xshift=2.6cm]

\begin{scope}[yshift=-0.35cm,xscale=1.7,xshift=-0.5cm,semithick]
\draw (0,3pt) -- (0,0) node [below] {\footnotesize $10$}
-- (1,0) node [below] {\footnotesize $40$} -- ++(0,3pt);
\foreach \x in {0.2, 0.4, 0.6, 0.8}
{
  \draw (\x,0) -- ++(0,3pt);
}
\end{scope}

\draw (0,-1.6) node {\small Strength\vphantom{Sg}};
\end{scope}

\begin{scope}[xshift=5.2cm]

\begin{scope}[yshift=-0.35cm,xscale=1.7,xshift=-0.5cm,semithick]
\draw (0,3pt) -- (0,0) node [below] {\footnotesize $0.2$}
-- (1,0) node [below] {\footnotesize $0.8$} -- ++(0,3pt);
\foreach \x in {0.333, 0.667}
{
  \draw (\x,0) -- ++(0,3pt);
}
\end{scope}

\draw (0,-1.6) node {\small Coherence\vphantom{Sg}};
\end{scope}

\end{tikzpicture}%
\endpgfgraphicnamed}
\caption{\label{fig:quantization}Typical structure tensor feature quantization
for BLADE filter selection, using 16 orientations, 5 strength bins, and 3
coherence bins.}
\end{figure}
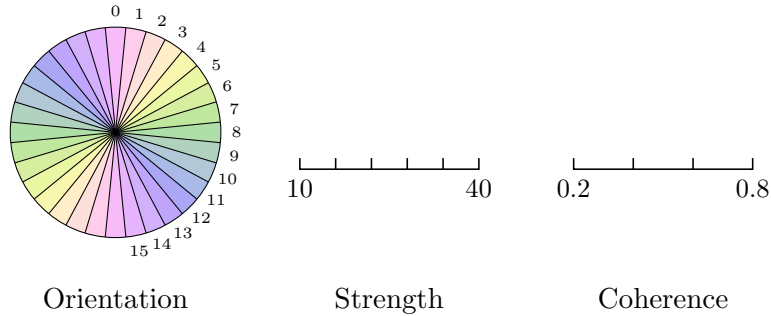

\subsection{Quantization}

We use the three described structure tensor features for filter selection $s$,
bounding and uniformly quantizing each feature to a small number of equal-sized
bins, and considering them together as a three-dimensional index into the bank
of filters.

A typical quantization is shown in Figure~\ref{fig:quantization}. The
orientation feature is quantized to 16 orientations. To avoid asymmetric
behavior, orientation quantization is done such that horizontal and vertical
orientations correspond to bin centers. Strength is bounded to $[10, 40]$ and
quantized to 5 bins, and coherence is bounded to $[0.2, 0.8]$ and quantized to
3 bins.

\section{Learning for Computational Photography}
\label{e:learning_image_processing_filters}

This and the next few sections show the flexibility of our approach by applying
BLADE to several applications. We begin by
demonstrating how BLADE can be used to make fast
approximations to other image processing methods. BLADE can
produce a similar effect that in some cases has lower computational cost
than the original method.

\begin{figure}[t]
\centering
\mbox{%
\beginpgfgraphicnamed{images/bilateral_filters}%
\input{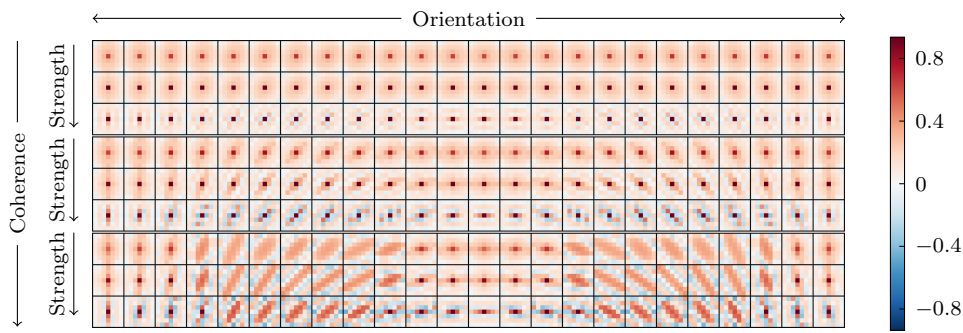}%
\endpgfgraphicnamed}
\caption{\label{fig:bilateral_filters} Trained $7\times 7$ BLADE filters
approximating the bilateral filter with 24 different orientations, 3 strength
values, and 3 coherence values.}
\end{figure}

\begin{figure}[t]
\centering
\mbox{%
\beginpgfgraphicnamed{images/bilateral_examples}%
\input{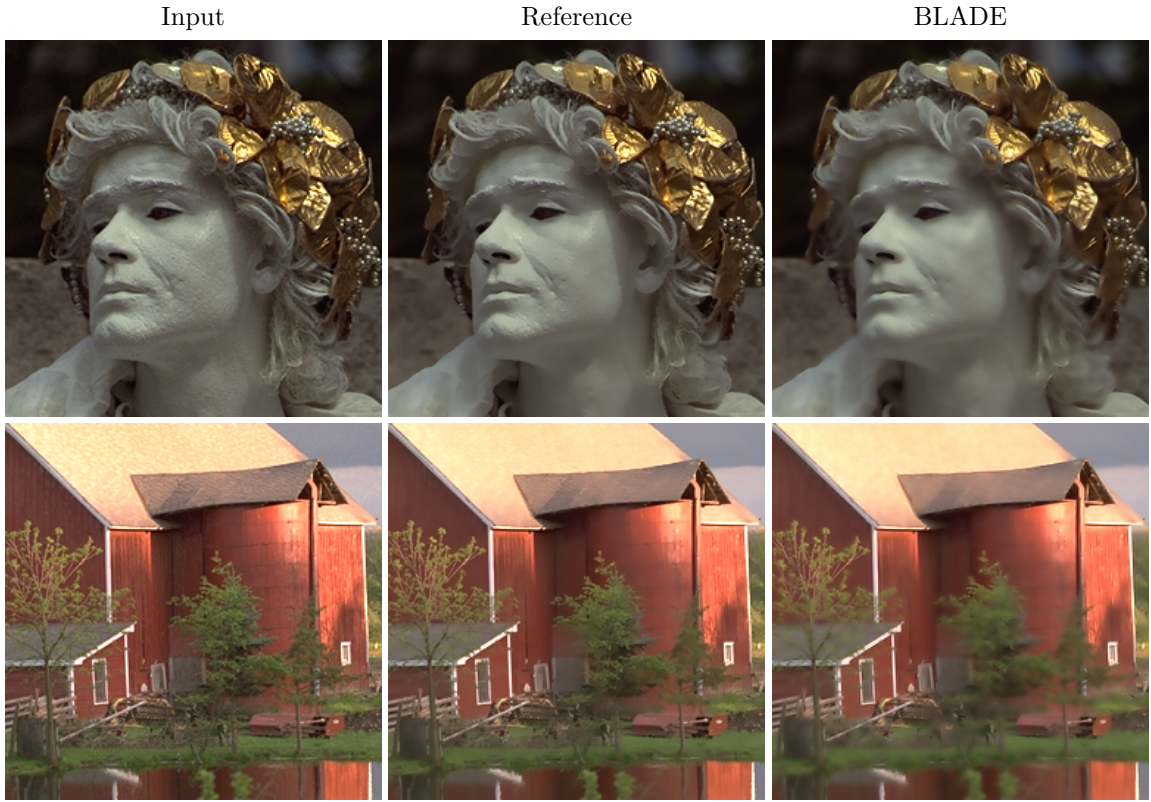}%
\endpgfgraphicnamed}
\caption{\label{fig:bilateral_examples} Approximation of the bilateral filter.
Top row: BLADE has PSNR 35.90\units{dB} and MSSIM 0.9272. Bottom row: BLADE has
PSNR 33.41\units{dB} and MSSIM 0.8735.}
\end{figure}

\begin{figure}[t]
\centering
\mbox{%
\beginpgfgraphicnamed{images/example_failed_training}%
\input{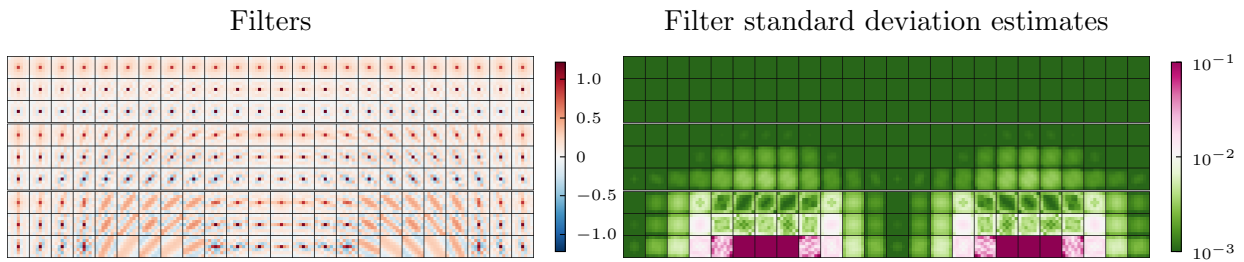}%
\endpgfgraphicnamed}
\caption{\label{fig:example_failed_training} Example of interpreting failed
training. Left: filters. Right: filter standard deviation estimates. Compare
with Figure~\ref{fig:bilateral_filters}.}
\end{figure}

\subsection{Bilateral filter}

The bilateral filter~\cite{tomasi1998bilateral} is an edge-adaptive smoothing
filter where each output pixel is computed from the input as
\begin{equation}
\outimage_i = \frac{\sum_{j}
G_{\sigma_r}(|\inimage_i - \inimage_j|) G_{\sigma_s}(\|i - j\|)
\inimage_j}{\sum_j G_{\sigma_r}(|\inimage_i - \inimage_j|)
G_{\sigma_s}(\|i - j\|)},
\end{equation}
where $G_\sigma$ denotes a Gaussian kernel with standard deviation $\sigma$.

We approximate the bilateral filter where the range kernel has standard
deviation $\sigma_r = 25$ (relative to a $[0, 255]$ nominal intensity range) and
the spatial kernel has standard deviation $\sigma_s = 2.5$ pixels. We use
$7\times 7$ filters, smoothing strength $\rho = 1.2$, 24 orientation buckets, 3
strength buckets over $[10, 35]$, and 3 coherence buckets over $[0.2, 0.8]$
(Fig.~\ref{fig:bilateral_filters}).

Over the Kodak Image Suite, bilateral BLADE agrees with the
reference bilateral implementation with an average PSNR of 37.30\units{dB} and
average MSSIM of 0.9630 and processing each $768\times 512$ image costs
18.6\units{ms} on a Xeon E5-1650v3 desktop PC. For comparison, the domain
transform by Gastal and Oliveira~\cite{gastal2011domain}, specifically
developed for efficiently approximating the bilateral filter, agrees with the
reference bilateral implementation with an average PSNR of 42.90\units{dB} and
average MSSIM of 0.9855 and costs 5.6\units{ms}.
Fig.~\ref{fig:bilateral_examples} shows BLADE approximation of the bilateral
filter on a crop from image 17 and 22 of the Kodak Image Suite.

\paragraph{Number of filters}
With using fewer filters, the BLADE approximation can be made to trade memory
cost for accuracy, which may be attractive on resource constrained platforms.
For example, training BLADE using instead 8 orientations, 3 strength buckets,
and no bucketing over coherence (24 filters vs.\ 216 filters above), the
approximation accuracy is only moderately reduced to an average PSNR of
37.00\units{dB} and MSSIM 0.9609. A reasonable approximation can be made with a
small number of filters.

\begin{figure}[t]
\centering
\mbox{%
\beginpgfgraphicnamed{images/tv_flow_filters}%
\input{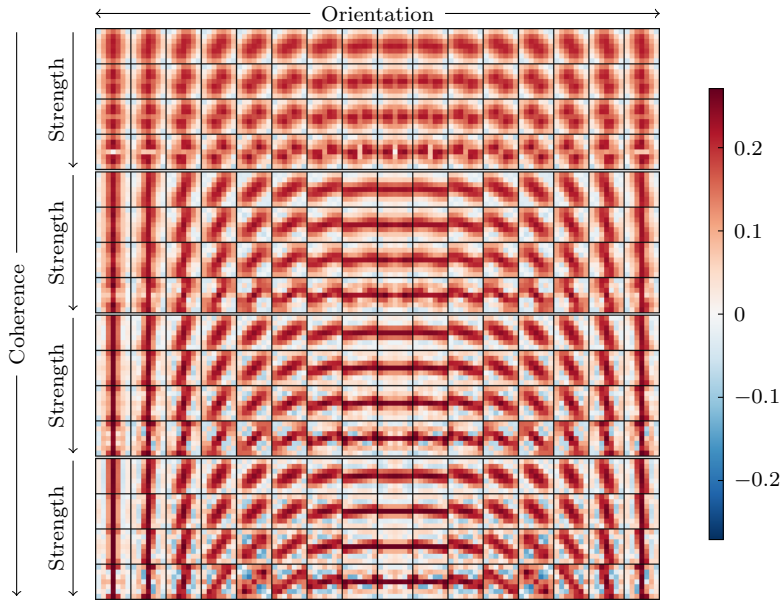}%
\endpgfgraphicnamed}
\caption{\label{fig:tv_flow_filters} $7\times 7$ BLADE filters approximating
TV flow with 16 different orientations, 4 strength values, and 4 coherence
values.}
\end{figure}

\begin{figure}[t]
\centering
\mbox{%
\beginpgfgraphicnamed{images/tv_flow_examples}%
\input{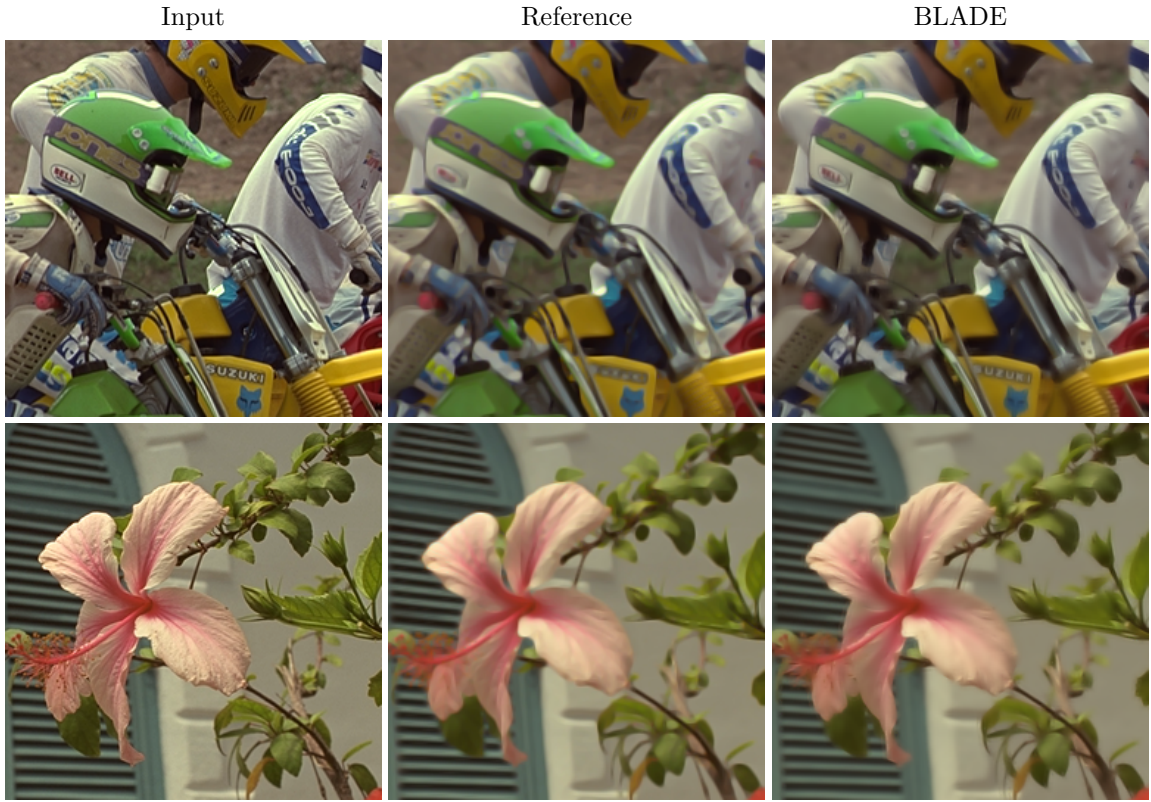}%
\endpgfgraphicnamed}
\caption{\label{fig:tv_flow_examples} Example of approximated TV flow. Top row:
BLADE has PSNR 32.10\units{dB} and MSSIM 0.9369. Bottom row: BLADE has PSNR
35.99\units{dB} and MSSIM 0.9691.}
\end{figure}

\paragraph{Interpretability}
To demonstrate the interpretability of BLADE,
Figure~\ref{fig:example_failed_training} shows a failed training example where
we attempted to train BLADE filters for bilateral filtering with strength range
over $[10, 80]$. Some of the high strength, high coherence buckets received few
training patches. Training parameters are otherwise the same as before. The
filters are overly-smooth in the problematic buckets. Additionally, the
corresponding filter standard deviation estimates are large, indicating a
training problem.

\subsection{TV flow}

In this section, we use BLADE to approximate the evolution of an
anisotropic diffusion equation, a modification of total variation (TV) flow
suggested by Marquina and Osher~\cite{marquina2000explicit},
\begin{equation}
\partial_t u = |\grad u| \div(\grad u /|\grad u|)
\end{equation}
where $u(x)$ is a continuous-domain image, $\grad$ denotes spatial gradient,
$\div$ denotes divergence, and $\partial_t u$ is the rate of change of the
evolution.

We train $7\times 7$ filters ($K = 16\times 4 \times 4$) on a dataset of 32
photograph images of size $1600\times 1200$. We use the second-order finite
difference scheme described by Marquina and Osher~\cite{marquina2000explicit} as
the reference implementation to generate target images for training
(Fig.~\ref{fig:tv_flow_filters}).

\begin{figure}[t]
\centering
\mbox{%
\beginpgfgraphicnamed{images/etf_filters}%
\input{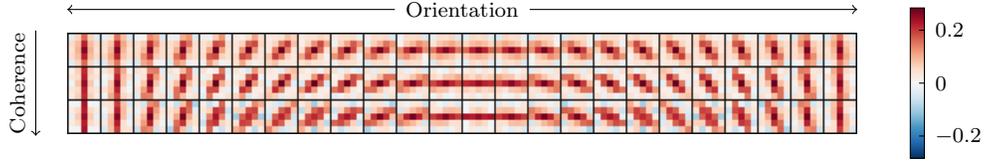}%
\endpgfgraphicnamed}
\caption{\label{fig:etf_filters} $5\times 5$ BLADE filters approximating edge
tangent flow with 24 different orientations and 3 coherence values.}
\end{figure}

\begin{figure}[t]
\centering
\mbox{%
\beginpgfgraphicnamed{images/etf_examples}%
\input{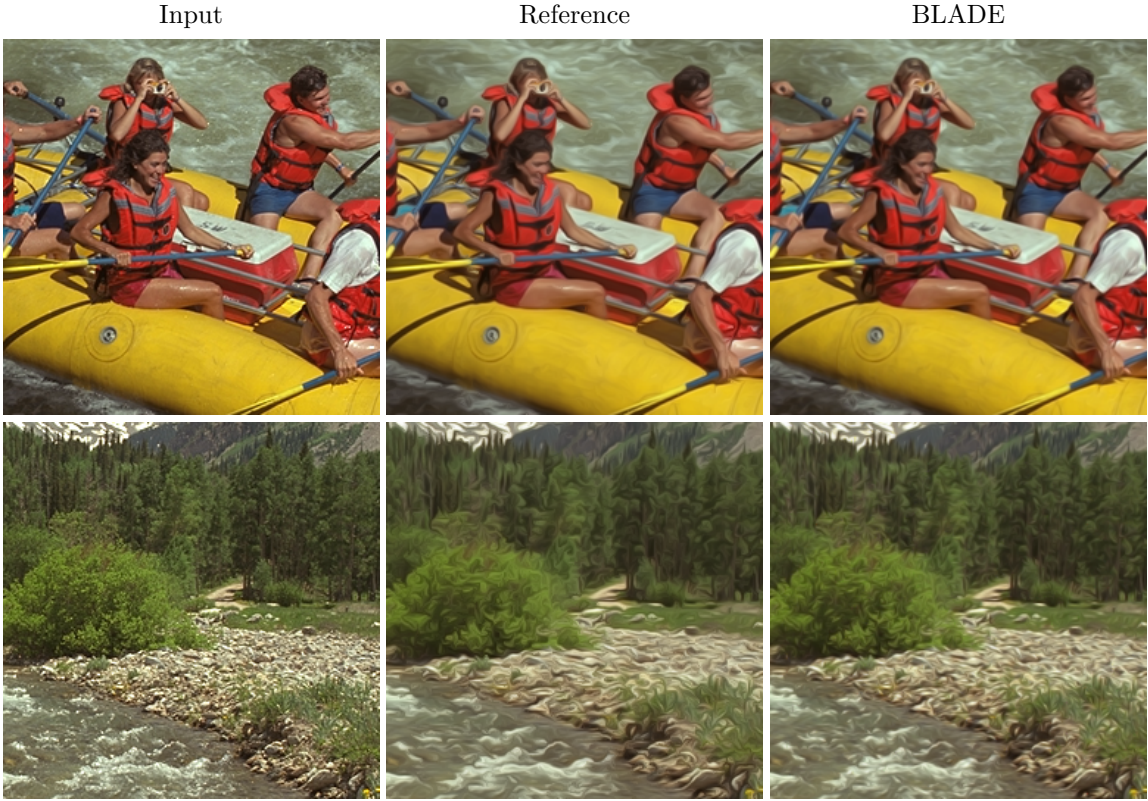}%
\endpgfgraphicnamed}
\caption{\label{fig:etf_examples} Examples of approximated edge tangent flow.
Top row: BLADE has PSNR 39.12\units{dB} and
MSSIM 0.9809. Bottom row: BLADE has PSNR 34.80\units{dB} and MSSIM 0.9639.}
\end{figure}

Over the Kodak Image Suite, the BLADE TV flow approximation agrees with the
reference implementation with an average PSNR of 34.77\units{dB} and average
MSSIM of 0.9482. Processing each $768\times 512$ image costs 17.4\units{ms} on a
Xeon E5-1650v3 desktop PC.
Fig.~\ref{fig:tv_flow_examples} shows the filters applied to a crop from image 5
and 8 of the Kodak Image Suite.

\subsection{Edge tangent flow}

Similarly, we can approximate tensor-driven diffusions like edge tangent flow
(ETF),
\begin{equation}\label{e:etf_pde}
\partial_t u = \div\bigl(D(u) \grad u\bigr)
\end{equation}
where at each point, $D(u)(x)$ is the $2\times 2$ outer product of the
unit-magnitude local edge tangent orientation. Supposing the edge tangent
orientation is everywhere equal to $\theta$, the diffusion (\ref{e:etf_pde})
reduces to the one-dimensional heat equation along $\theta$, whose solution is
convolution with an oriented Gaussian,
\begin{equation}
u(t,x) = \int_{-\infty}^\infty
\frac{\exp(-\tfrac{s^2}{4t})}{\sqrt{4\pi t}} u(0, x + \theta s)\,ds.
\end{equation}
We expect for this reason that the structure tensor orientation matters
primarily to approximate this diffusion. However, if orientation is not locally
constant, the solution is more complicated. Therefore, we use also the
coherence, which indicates the extent to which the constant orientation
assumption is true (whereas strength gives no such indication, so we exclude
it).

We train $5\times 5$ filters ($K = 24 \times 3$) over the same dataset, using
line integral convolution evolved with second-order Runge--Kutta as a reference
implementation to generate target images for training.

Over the Kodak Image Suite, the BLADE ETF approximation agrees with the
reference implementation with an average PSNR of 40.94\units{dB} and
average MSSIM of 0.9849. Processing each $768\times 512$ image costs
14.5\units{ms} on a Xeon E5-1650v3 desktop PC.
Figure~\ref{fig:etf_examples} shows an example application to a crop from
image 13 and 14 of the Kodak Image Suite.

\begin{figure}[t]
\centering
\includegraphics[width=0.95\textwidth]{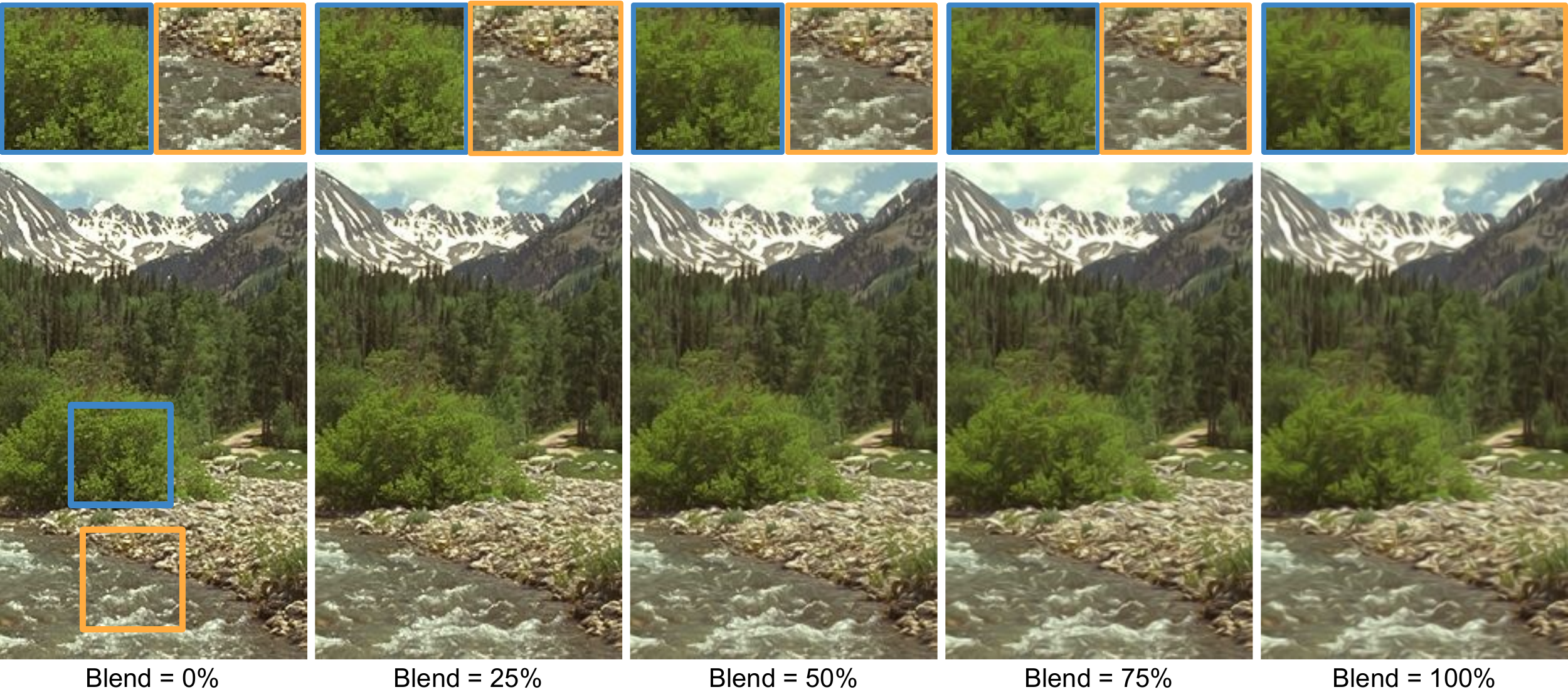}
\caption{\label{fig:blend_filter_with_identity}Control filter strength. Blending
the identity filter with the edge tangent flow filter.}
\end{figure}

\begin{figure}[t]
\centering
\mbox{%
\beginpgfgraphicnamed{images/simple_denoiser_filters}%
\input{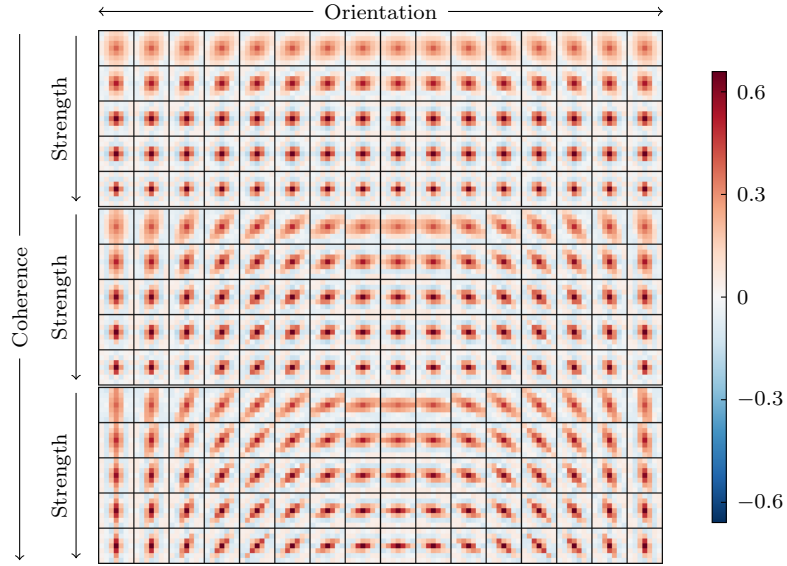}%
\endpgfgraphicnamed}
\caption{\label{fig:simple_denoiser_filters} $7\times 7$ BLADE filters for AWGN
denoising with noise standard deviation 20, using $\rho = 1.7$, 16 different
orientations, 5 strength values, and 3 coherence values.}
\end{figure}

\begin{figure}[t]
\centering
\mbox{%
\beginpgfgraphicnamed{images/simple_denoiser_kodak_05}%
\input{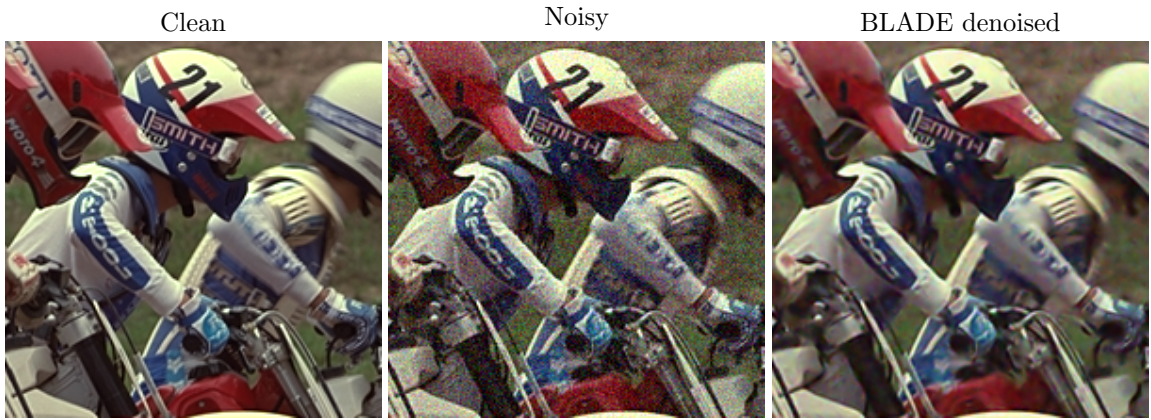}%
\endpgfgraphicnamed}
\caption{\label{fig:simple_denoiser_kodak_05} Simple AWGN denoiser example
with noise standard deviation 20. The noisy image has PSNR 22.34\units{dB}
and MSSIM 0.5792 and the BLADE denoised image has PSNR 28.04\units{dB} and
MSSIM 0.8268.}
\end{figure}

\subsection{Control the filter strength}

One limitation of using these filters for image processing is that each set of
filters is trained for a specific set of filter parameters (e.g.\ our bilateral
filter is trained for  $\sigma_r = 25$, $\sigma_s = 2.5$),
however, it would be interesting to be able to control the strength of the
effect without having multiple versions of the filters for different filter
parameters. To address this issue, given that each filter knows how to apply the
effect for that specific bucket, we interpolate each filter linearly with the
identity filter (i.e., a delta in the origin of the filter).
Figure~\ref{fig:blend_filter_with_identity} shows an example of how we can
use this to control the strength of the BLADE ETF filter.

Since we interpolate the filters (which are small compared to the 
image), the performance penalty is negligible.

\section{Denoising}\label{sec:denoising}

\subsection{A Simple AWGN Denoiser}\label{sec:a_simple_awgn_denoiser}

We build a grayscale image denoiser with an additive white Gaussian
noise (AWGN) model by training a BLADE filter,
using a set of clean images as the targets and synthetically adding noise to
create the observations.

Filters are selected based on the observed image structure tensor analysis (as
described in section~\ref{sec:filter_selection}). Since the observed image is
noisy, the structure tensor smoothing by $G_\rho$ has the critical role of
ameliorating noise before computing structure tensor features. Parameter $\rho$
must be large enough for the noise level to obtain robust filter selection.
Figure~\ref{fig:simple_denoiser_filters} shows the trained filters for a noise
standard deviation of 20, for which we set $\rho = 1.7$.

We test the denoiser over the Kodak Image Suite with ten AWGN noise realizations
per image. The noisy input images have average PSNR of 22.31\units{dB} and
average MSSIM of 0.4030. The simple BLADE denoiser improves the average PSNR to
29.44\units{dB} and MSSIM to 0.7732. Processing each $768\times 512$ image costs
19.8\units{ms} on a Xeon E5-1650v3 desktop PC.
Figure~\ref{fig:simple_denoiser_kodak_05} shows an example of denoising a crop
of image 5 from the Kodak Image Suite.


\begin{figure}[t]
\centering
\begin{minipage}[b]{0.62\textwidth}
  \includegraphics[width=\textwidth]{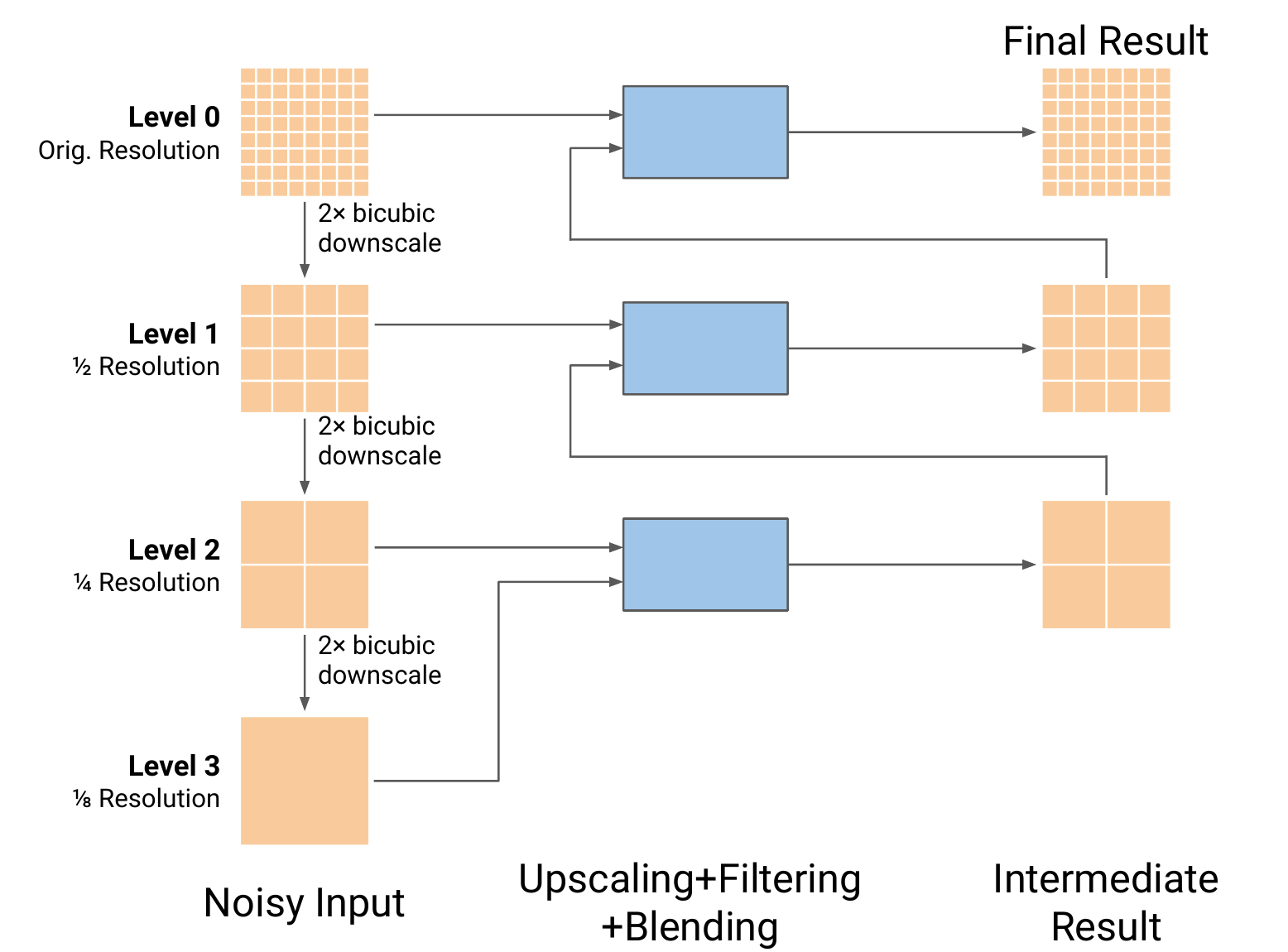}
\end{minipage}
\hfill
\begin{minipage}[b]{0.28\textwidth}
  \includegraphics[width=\textwidth]{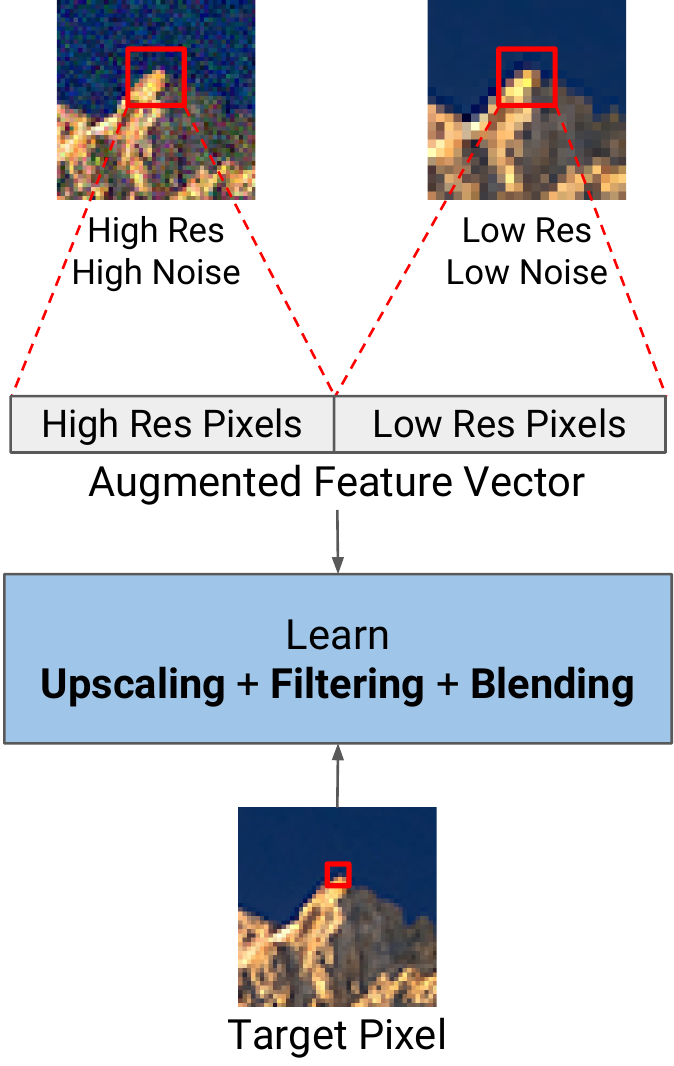}
\end{minipage}
\caption{\label{fig:multiscale_denoising_pipeline} Multilevel BLADE denoising
pipeline.}
\end{figure}

\begin{figure}[t]
\centering
\mbox{%
\beginpgfgraphicnamed{images/multiscale_denoising}%
\input{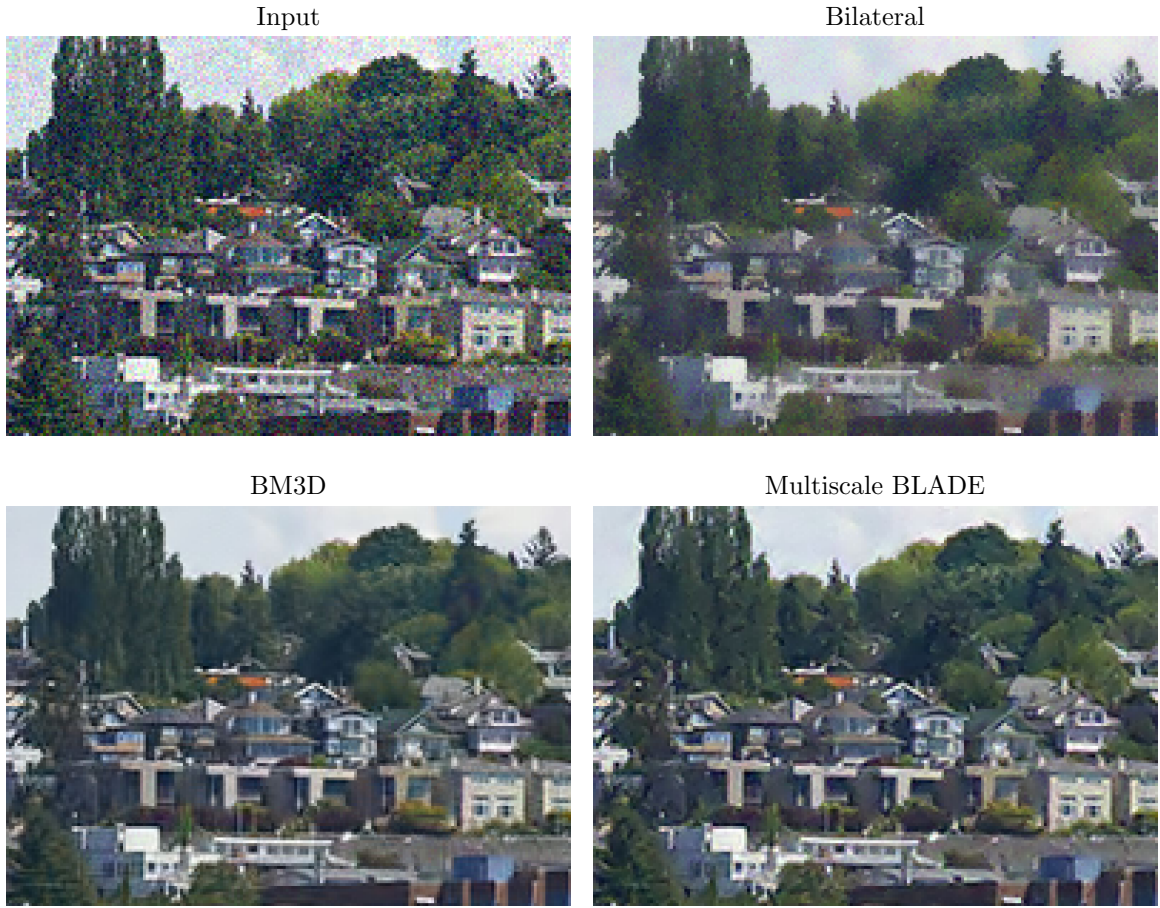}%
\endpgfgraphicnamed}
\caption{\label{fig:multiscale_denoising} Qualitative comparison of
denoising results.}
\end{figure}

\subsection{Multiscale AWGN Denoising}\label{sec:multiscale_awgn_denoising}

We now demonstrate how multiscale application of the previous section enables
BLADE to perform fast edge-adaptive image denoising with quality in the ballpark
of much more expensive methods. We begin by taking an input image and forming an
image pyramid by downsampling by factor of two. Using a bicubic downsampler,
which effectively reduce the noise level by half per level, and is extremely
fast. An image pyramid of a target image is also constructed by using the same
downsampling method.

Considering level $L$ as the coarsest level of the image pyramid, we start
training from the level $L-1$ and go up to the finest level. We train filters
that operate on a pair of patches, one from the input image at the current level
and another at the corresponding position in the filtered result at the next
coarser level, see Figure~\ref{fig:multiscale_denoising_pipeline}.

We build up denoised results as shown in
Figure~\ref{fig:multiscale_denoising_pipeline}. The filters are learned to
optimally upscale coarser level images and blend them into filtered current
level images to create results for next level. Structure tensor analysis is
performed on the patches of the current level.

Figure~\ref{fig:multiscale_denoising} shows a comparison of denoising
results among other methods. Multiscale denoising results have quality
comparable to the state-of-the-art algorithms but with fast processing time.

\begin{figure}[!h]
\centering
\mbox{%
\beginpgfgraphicnamed{images/unjpeg_filters}%
\input{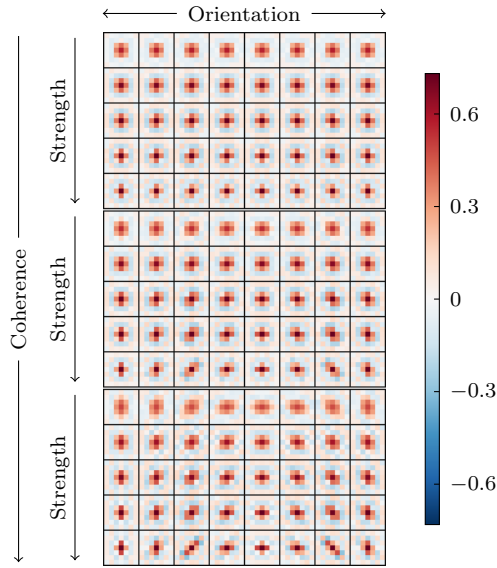}%
\endpgfgraphicnamed}
\caption{\label{fig:unjpeg_filters} $7\times 7$ BLADE filters for JPEG
compression artifact removal for JPEG quality level 50, using $\rho = 1.2$, 8
different orientations, 5 strength values, and 3 coherence values.}
\end{figure}

\begin{figure}[!h]
\centering
\mbox{%
\beginpgfgraphicnamed{images/unjpeg_kodak_21}%
\input{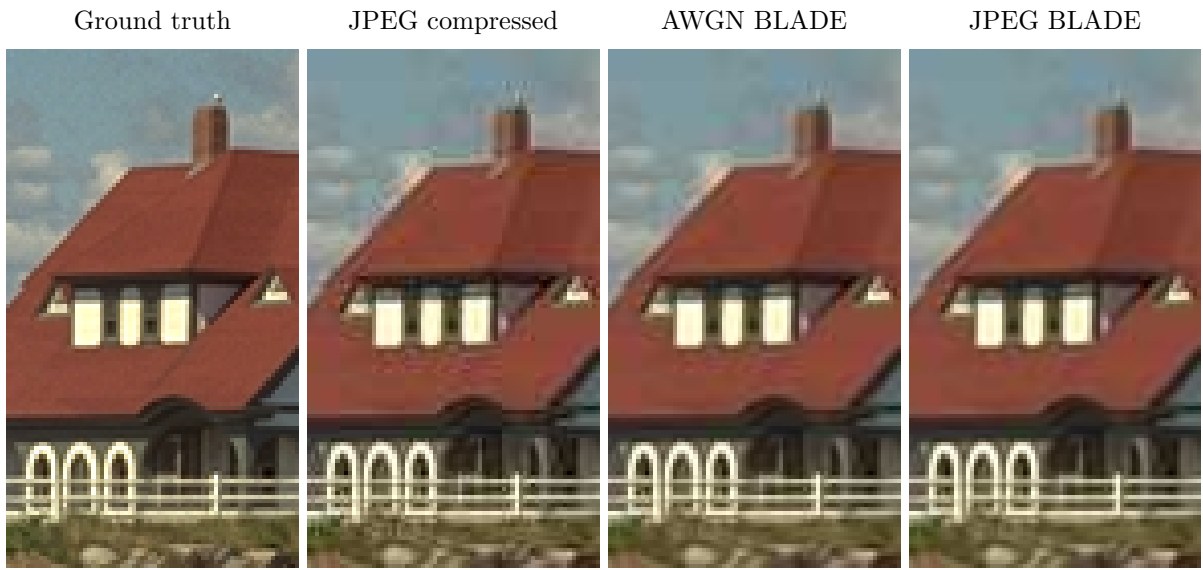}%
\endpgfgraphicnamed}
\caption{\label{fig:unjpeg_kodak_21} JPEG artifact removal with BLADE for JPEG
quality level 50.}
\end{figure}

\section{JPEG Artifact Removal}\label{sec:jpeg_artifact_removal}

Besides AWGN denoising, BLADE filters can undo other kinds of distortion such as
JPEG compression artifacts. We train a BLADE filter, using a set of clean images
as the targets and JPEG-compressed versions of the same images with JPEG quality
level 50. For simplicity, we train on the luma channel only. At inference time,
we use the luma channel for filter selection, then process each RGB color
channel independently with those selected filters. A more complicated scheme
could train other sets of filters for the $\mathrm{C_b}$ and $\mathrm{C_r}$
channels.

Fig.~\ref{fig:unjpeg_kodak_21} shows the filters applied to a crop from image 21
of the Kodak Image Suite. Filtering greatly reduces the visibility of the
$8\times 8$ block edges and DCT ripples.

JPEG artifact removal is essentially a denoising problem, viewing distortion
introduced by the lossy compression as noise. We measure over the Kodak Image
Suite that JPEG compression with quality 50 corresponds to an average MSE of
43.6. For comparison, we show the results of the simple AWGN BLADE denoiser from
section~\ref{sec:a_simple_awgn_denoiser} trained for AWGN noise of variance 43.6
(``AWGN BLADE'' in Fig.~\ref{fig:unjpeg_kodak_21}). While the AWGN denoiser
reduces much of the artifacts, the BLADE trained on JPEG is more effective.

Over the Kodak Image Suite, quality level 50 JPEG compression has average PSNR
32.17\units{dB}. AWGN BLADE improves the average PSNR to 32.66\units{dB}, while
JPEG BLADE improves average PSNR to 32.75\units{dB}. Indeed, JPEG noise is not
AWGN. This shows BLADE takes advantage of spatial correlations in the noise for
which it is trained.

Alternatively, we can account for JPEG's $8\times 8$ block structure by training
different filters per pixel coordinate modulo 8. At the memory cost of 64 times
more filters compared to the single shift approach above, this extension makes a
modest improvement.

\begin{figure}[t]
\centering
\mbox{%
\includegraphics[width=10cm]{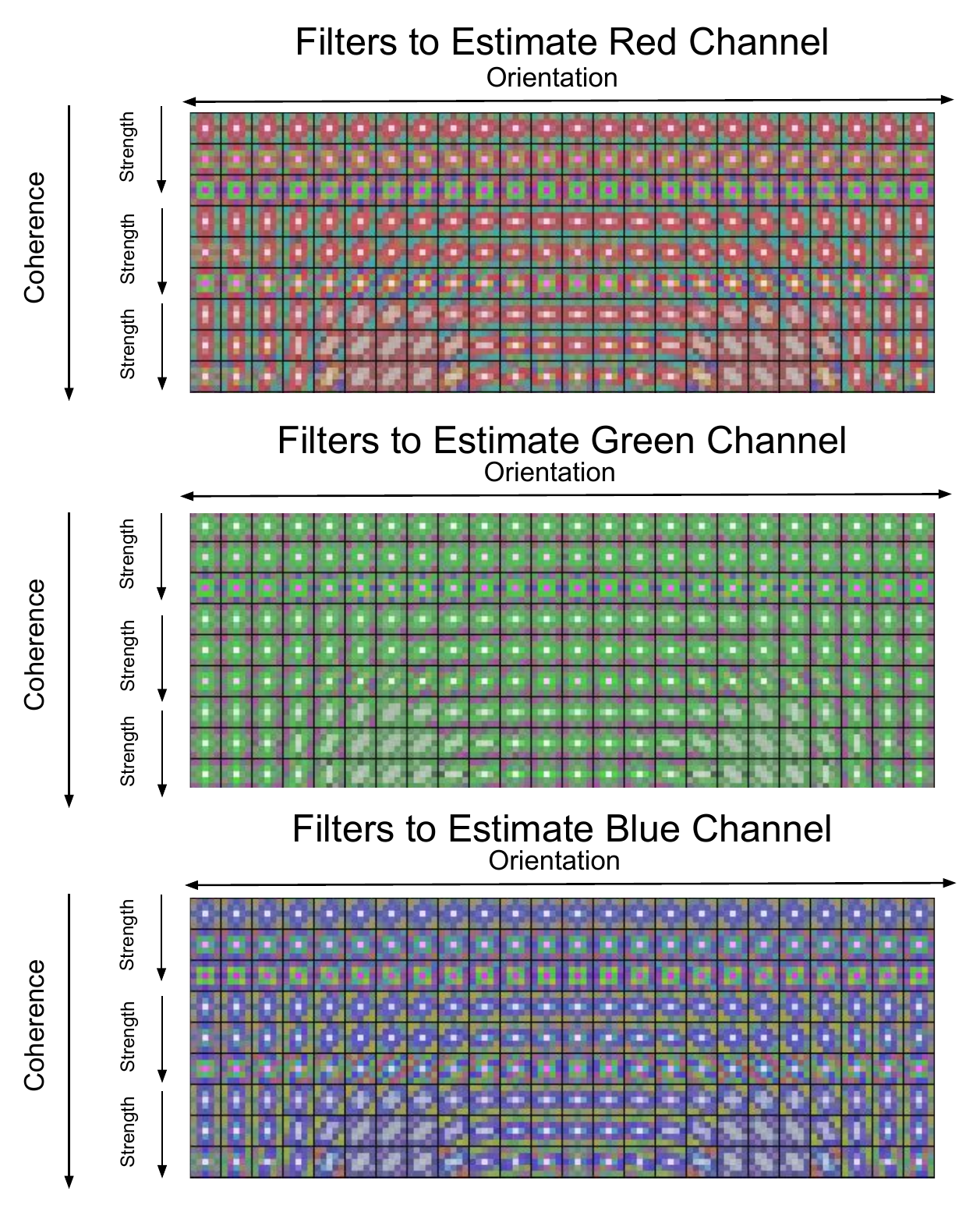}}
\caption{\label{fig:color_raisr_filters} Color BLADE filters trained on Menon
demosaiced images, with $\rho=0.7$, 8 difference orientations, 3 strength
values, and 3 coherence values. Here, the color map uses the red, green, and
blue components of the filters.}
\end{figure}

\begin{figure}[t]
\centering
\mbox{%
\includegraphics[width=0.95\textwidth]{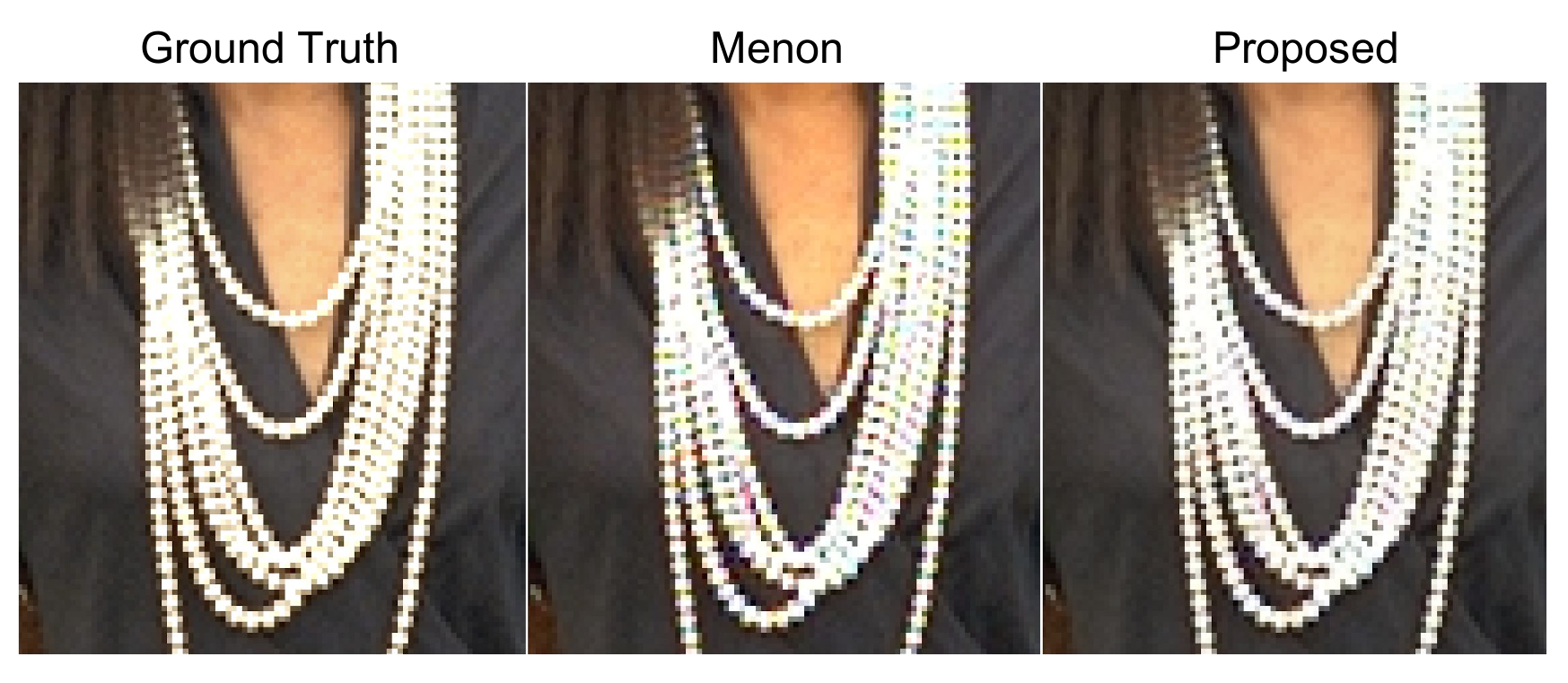}}
\caption{\label{fig:demosaic_example}Comparison between Menon and BLADE. 
Demosaicing with BLADE displays less artifacts.}
\end{figure}

\section{Demosaicing}\label{sec:demosaicing}

In this section, we consider the task of demosaicing using BLADE. In the
original RAISR formulation, filters are designed only for single-channel images,
and operate on the luma channel for color images. Demosaicing requires
exploiting correlation between color channels, which we do by computing each
output sample as a linear combination of samples from all three channels in the
input patch.

For each pixel $i$, instead of a single-channel filter $\vv{\filter}^{s(i)}$
that takes in a single-channel patch and outputs a pixel, we use three filters
$\vv{\filter}^{s(i), r}$, $\vv{\filter}^{s(i), g}$, and $\vv{\filter}^{s(i), b}$
that compute output red, green, and blue pixels respectively from a given RGB
input patch. The resulting inference becomes:
\begin{equation}
\begin{aligned}
\outimage_i^r &= (\vv{\filter}^{s(i), r})^T \vv{R}_i \vv{\inimage}\\
\outimage_i^g &= (\vv{\filter}^{s(i), g})^T \vv{R}_i \vv{\inimage}\\
\outimage_i^b &= (\vv{\filter}^{s(i), b})^T \vv{R}_i \vv{\inimage}
\end{aligned}
\end{equation}
where $\vv{R}_i$ extracts a color patch centered at $i$, and $\outimage_i^r$,
$\outimage_i^g$, and $\outimage_i^b$ denote the output red, green, and blue
pixels. We interpret this extension as three separate BLADE
filterbanks described in Section~\ref{sec:blade}, one for each output color
channel.

With this color extension, we then train filters to exploit correlations between
color channels for demosaicing. Similar to RAISR upscaling, we first apply a
fast cheap demosaicing on the input image, then perform structure tensor
analysis and filter selection as usual. For our experiments we use the method
described in Menon et al.~\cite{menon2007demosaicing}.

Figure~\ref{fig:color_raisr_filters} shows the color filters trained on Bayer
demosaiced images using the method from Menon et al.~\cite{menon2007demosaicing}
on the Kodak Image Suite, with $\rho=0.7$, 8 different orientations, 3 strength
values, and 3 coherence values. Three sets of color filters were trained to
predict red, green, and blue pixels respectively. As expected, the filters to
estimate each color channel mostly utilize information from the same color
channel. For example, the color filters to estimate red channel are mostly red.
On the other hand, cross-channel correlation is leveraged as the filters are not
purely red, green, or blue.

We evaluate the demosaicing methods on the Kodak Image Suite. The average PSNR
for Menon demosaiced images is 39.14\units{dB}, whereas the average PSNR for our
proposed method is 39.69\units{dB}. Figure~\ref{fig:demosaic_example} shows a
cropped example with reduction of demosaicing artifacts using our proposed
method.

\begin{figure}[t]
\centering%
\mbox{%
\beginpgfgraphicnamed{images/cpu_time_vs_size}%

\begin{tikzpicture}[xscale=0.1,yscale=1.0,scale=1.25]

\pgfdeclareplotmark{o}
{
  \pgftransformresetnontranslations
  \pgftransformscale{0.025} 
  \fill (0,0) circle (1.16);
}

\draw (0,4)
-- (0,0) node [midway,rotate=90,above=16pt,black] {\small CPU time (s)}
-- (64,0) node [midway,below=12pt,black] {\small Image size (MP)};

\foreach \x in {1, 4, 16, 64}
{
  \begin{scope}[xshift=\x cm]
  \pgftransformresetnontranslations
  \draw (0,2pt) -- (0,0) node [below=-1pt,black] {\footnotesize $\x$};
  \end{scope}
}
\foreach \y in {0, 2, 4}
{
  \begin{scope}[yshift=\y cm]
  \pgftransformresetnontranslations
  \draw (2pt,0) -- (0,0) node [left,black] {\footnotesize $\y$};
  \end{scope}
}

\draw plot [mark=o,smooth] coordinates
{(1,0.092) (4,0.303) (16,1.186) (64,4.564)}
node [left,yshift=2pt] {\scriptsize $13\times 13$};
\draw plot [mark=o,smooth] coordinates
{(1,0.046) (4,0.157) (16,0.599) (64,2.320)}
node [left,yshift=2pt] {\scriptsize $11\times 11$};
\draw plot [mark=o,smooth] coordinates
{(1,0.027) (4,0.103) (16,0.396) (64,1.546)}
node [left,yshift=2pt] {\scriptsize $9\times 9$};
\draw plot [mark=o,smooth] coordinates
{(1,0.017) (4,0.065) (16,0.247) (64,0.986)}
node [left,yshift=2.5pt] {\scriptsize $7\times 7$};
\draw plot [mark=o,smooth] coordinates
{(1,0.011) (4,0.043) (16,0.166) (64,0.672)}
node [left,yshift=1.7pt] {\scriptsize $5\times 5$};

\end{tikzpicture}%
\endpgfgraphicnamed}
\caption{\label{fig:performance_vs_image_size}CPU time on Xeon E5-1650v3 desktop
PC vs.\ image size in megapixels.}
\end{figure}
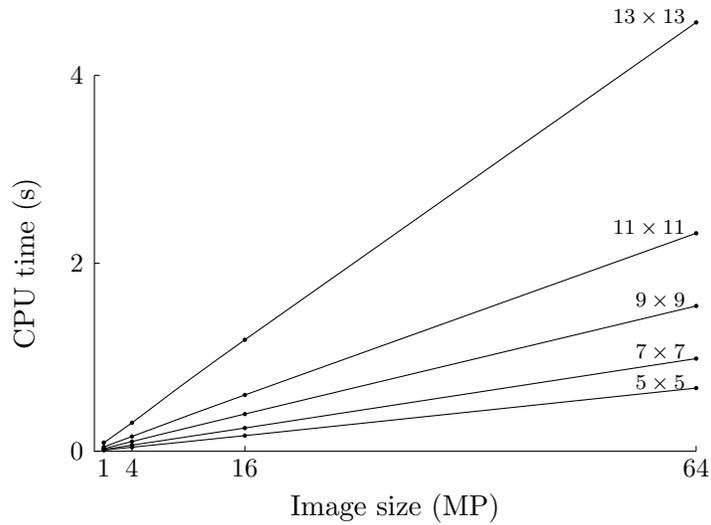

\section{Computational Performance}\label{sec:computational_performance}

BLADE filtering has been optimized to run fast on desktop and mobile
platforms. The optimizations to achieve the desired performance are the
following:
\begin{itemize}
\item For CPU implementation, the code is optimized using the programming
language Halide~\cite{ragan2012decoupling}. Halide code decouples the algorithm
implementation from the scheduling. This allows us to create optimized code
that uses native vector instructions and parallelizes over multiple
CPU cores.
\item BLADE processing on the CPU is performed using low-precision integer
arithmetic where possible. This allows for a higher degree of vectorization,
especially on mobile processors, and reduces memory bandwidth.
Analogously, the GPU implementation uses 16-bit float arithmetic where
possible.
\item The algorithm is GPU amenable, we have seen up to an order of magnitude 
performance improvement on GPU over our CPU implementation. The BLADE filter 
selection maps
efficiently to texture fetches with all filter coefficients stored in a single
RGBA texture. It is also possible to process 4 pixels in parallel per pixel
shader invocation by taking advantage of all 4 RGBA channels for processing.
\item Approximations to transcendental functions are used where applicable.
For example the arctangent for orientation angle computation uses a variation of
a well-known quadratic approximation~\cite{rajan06}.
\end{itemize}


Our algorithm has runtime linear in the number of output pixels.
Figure~\ref{fig:performance_vs_image_size} shows that given a fixed filter size,
the performance is linear.

Table~\ref{table:performance_vs_device} shows the performance of BLADE for
different platforms on the CPU and different filter sizes. A full HD image
($1920\times1080$ pixels) takes less than 60\units{ms} to process on the mobile
devices we tested.

\begin{table}[t]
\centering
\caption{\label{table:performance_vs_device}
CPU performance on different devices.}
\begin{small}
\begin{tabular}{lccccc}
\hline
Platform & $5\times5$ filters & $7\times7$ filters & $9\times9$ filters & $11\times$ filters  & $13\times13$ filters  \\
\hline
Xeon E5-1650v3 PC&97.50\units{MP/s} &71.70\units{MP/s} &52.26\units{MP/s} &41.28\units{MP/s} &31.00\units{MP/s} \\
Pixel 2017&29.47\units{MP/s} &22.41\units{MP/s} &18.00\units{MP/s} &15.01\units{MP/s} &11.39\units{MP/s} \\
Pixel 2016&21.39\units{MP/s} &15.62\units{MP/s} &11.42\units{MP/s} &8.71\units{MP/s} &6.59\units{MP/s} \\
Nexus 6P&19.06\units{MP/s} &14.60\units{MP/s} &11.09\units{MP/s} &9.11\units{MP/s} &6.72\units{MP/s} \\
\hline
\end{tabular}
\end{small}
\end{table}


The GPU implementation focuses on the $5\times5$ filters, the peak performance
are as follows: $131.53\units{MP/s}$ on a Nexus 6P, $150.09\units{MP/s}$ on a
Pixel 2016, $223.03\units{MP/s}$ on a Pixel 2017.  Another way to look at it is
that the algorithm is capable of 4k video output ($3840\times2160$ pixels) at
27\units{fps} or at full HD output at over 100\units{fps} on device.

\section{Conclusions}\label{sec:conclusions}

We have presented Best Linear Adaptive Enhancement (BLADE), a framework for
simple, trainable, edge-adaptive filtering based on a local linear estimator.
BLADE has computationally efficient inference, is easy to train and interpret,
and is applicable to a fairly wide range of tasks. Filter selection is not
trained; it is performed by hand-crafted features derived from the image
structure tensor, which are effective for adapting behavior to the local image
geometry, but is probably the biggest weakness of our approach from a machine
learning perspective and an interesting aspect for future work.

\appendix

\section{Numerical details}

Implementation of (\ref{e:structure_tensor}) requires a numerical approximation
of the image gradient. Forward (or backward) differences could be used, but
they are only first-order accurate,
\begin{equation}
\tfrac{1}{h} \bigl(u(x_1 + 1, x_2) - u(x_1,x_2)\bigr) =
\partial_{x_1} u(x_1,x_2) + \tfrac{1}{2} \partial_{x_1^2}^2 u(x_1,x_2) h
+ O(h^2),
\end{equation}
where $h$ is the width of a pixel.
Centered differences are second-order accurate, but they increase the footprint
of the approximation
\begin{equation}
\tfrac{1}{2h} \bigl(u(x_1 + 1, x_2) - u(x_1 - 1,x_2)\bigr) =
\partial_{x_1} u(x_1,x_2) + \tfrac{1}{12} \partial_{x_1^3}^3 u(x_1,x_2) h^2
+ O(h^4)
\end{equation}

\begin{figure}[t]
\centering
\mbox{%
\beginpgfgraphicnamed{images/diagonal_diffs}%

\begin{tikzpicture}[scale=0.9,yscale=-1,>=stealth']

\begin{scope}
\draw [gray] (0,-0.7) node [above=-1.5pt] {\scriptsize $x_1$} -- ++(0,2.5pt)
(1,-0.7) node [above=-1.5pt] {\scriptsize $x_1+1$} -- ++(0,2.5pt)
(-0.7,0) node [left=-1.5pt] {\scriptsize $x_2$} -- ++(2.5pt,0)
(-0.7,1) node [left=-1.5pt] {\scriptsize $x_2+1$} -- ++(2.5pt,0)
(-0.25,-0.7) -- (1.25,-0.7) (-0.7,-0.1) -- (-0.7,1.1);

\node [draw,circle,inner sep=1pt] (p00) at (0,0) {};
\node [draw,circle,inner sep=1pt] (p10) at (1,0) {};
\node [draw,circle,inner sep=1pt] (p01) at (0,1) {};
\node [draw,circle,inner sep=1pt] (p11) at (1,1) {};

\draw [semithick] (p10) -- (p01);
\draw (p10) node [above] {\footnotesize $+1$};
\draw (p01) node [above,xshift=-0.4em] {\footnotesize $-1$};

\draw (0.5,1) node [below=8pt] {$\approx\partial_{x_1'}$};
\end{scope}

\begin{scope}[xshift=4.6cm]
\draw [gray] (0,-0.7) node [above=-1.5pt] {\scriptsize $x_1$} -- ++(0,2.5pt)
(1,-0.7) node [above=-1.5pt] {\scriptsize $x_1+1$} -- ++(0,2.5pt)
(-0.7,0) node [left=-1.5pt] {\scriptsize $x_2$} -- ++(2.5pt,0)
(-0.7,1) node [left=-1.5pt] {\scriptsize $x_2+1$} -- ++(2.5pt,0)
(-0.25,-0.7) -- (1.25,-0.7) (-0.7,-0.1) -- (-0.7,1.1);

\node [draw,circle,inner sep=1pt] (p00) at (0,0) {};
\node [draw,circle,inner sep=1pt] (p10) at (1,0) {};
\node [draw,circle,inner sep=1pt] (p01) at (0,1) {};
\node [draw,circle,inner sep=1pt] (p11) at (1,1) {};

\draw [semithick] (p11) -- (p00);
\draw (p11) node [above,xshift=0.4em] {\footnotesize $+1$};
\draw (p00) node [above] {\footnotesize $-1$};

\draw (0.5,1) node [below=8pt] {$\approx\partial_{x_2'}$};
\end{scope}

\end{tikzpicture}%
\endpgfgraphicnamed}
\caption{\label{fig:diagonal_diffs}Diagonal finite differences for approximating
the image gradient.}
\end{figure}
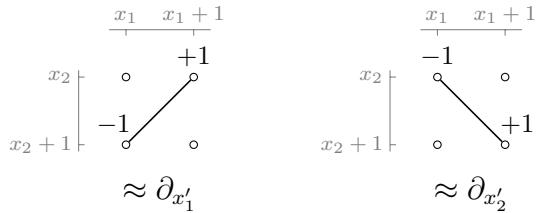

We instead approximate derivatives of $45^\circ$-rotated axes $x' =
\tfrac{1}{\sqrt{2}} \bigl(\begin{smallmatrix} 1 & -1 \\ 1 & 1
\end{smallmatrix}\bigr) x$ with diagonal differences as depicted in
Figure~\ref{fig:diagonal_diffs}. If interpreted as logically located at cell
centers, diagonal differences are second-order accurate,
\begin{align}
\tfrac{1}{\sqrt{2} h} \bigl(u(x_1 + 1, x_2) - u(x_1,x_2 + 1)\bigr) &=
\partial_{x_1'} u(x_1+\tfrac{1}{2},x_2+\tfrac{1}{2}) + O(h^2), \\
\tfrac{1}{\sqrt{2} h} \bigl(u(x_1 + 1, x_2 + 1) - u(x_1,x_2)\bigr) &=
\partial_{x_2'} u(x_1+\tfrac{1}{2},x_2+\tfrac{1}{2}) + O(h^2).
\end{align}
We then carry out structure tensor analysis on this $45^\circ$-rotated and
$1/2$-sample shifted gradient approximation. When smoothing with $G_\rho$, a
symmetric even-length FIR filter is used in each dimension to compensate for the
$+1/2$ shift. The eigenvector $\vv{w}$ is rotated back by $45^\circ$ to
compensate for the rotation, $\vv{\Tilde{w}} = \bigl(\begin{smallmatrix} 1 & 1
\\ -1 & 1 \end{smallmatrix}\bigr) \vv{w}$.

\bibliography{references}

\end{document}